\documentclass[conference]{IEEEtran}
\ifCLASSINFOpdf
\else
\fi
\usepackage{cite}
\usepackage{amsmath,amssymb,amsfonts}
\usepackage{balance}
\usepackage{graphicx}
\usepackage{textcomp}
\usepackage{xcolor}
\usepackage{todonotes}
\usepackage{bm,nicefrac}
\usepackage{cancel}

\usepackage{framed,url,caption,subcaption,graphicx}
\usepackage{amsmath,xcolor,soul, fontawesome,algpseudocode,algorithm,xspace,multirow,ctable,longtable,colortbl, lscape, pifont}
\usepackage{amssymb, cleveref}
\usepackage{booktabs, siunitx,comment}

\definecolor{light-gray}{gray}{0.9}
\definecolor{darkgreen}{rgb}{0,0.5,0}
\definecolor{light-blue}{rgb}{0,.7,1}
\definecolor{red}{rgb}{.7, 0, 0}
\let\labelindent\relax
\usepackage{enumitem}
\setlist{noitemsep, topsep=0pt, partopsep=0pt}
\graphicspath{{./figures/} }
\crefformat{section}{\S#2#1#3}
\crefformat{subsection}{\S#2#1#3}
\crefformat{subsubsection}{\S#2#1#3}

\expandafter\def\expandafter\UrlBreaks\expandafter{\UrlBreaks
  \do\a\do\b\do\c\do\d\do\e\do\f\do\g\do\h\do\i\do\j%
  \do\k\do\l\do\m\do\n\do\o\do\p\do\q\do\r\do\s\do\t%
  \do\u\do\v\do\w\do\x\do\y\do\z\do\A\do\B\do\C\do\D%
  \do\E\do\F\do\G\do\H\do\I\do\J\do\K\do\L\do\M\do\N%
  \do\O\do\P\do\Q\do\R\do\S\do\T\do\U\do\V\do\W\do\X%
  \do\Y\do\Z}

\newcommand{\F}{Figure }
\newcommand{\Fs}{Figures }

\newcommand{\T}{Table }

\renewcommand{\S}{Section }
\newcommand{\A}{Alg.}
\newcount\Comments  \Comments=1  
\newcommand{\kibitz}[2]{\ifnum\Comments=1\textcolor{#1}{#2}\fi}
\newcommand{\STOP}[1]  {\kibitz{red}   {[\textbf{STOP HERE} - compiled: \today]}}

\newcommand{\R}{\mathbb{R}}

\newcommand{\paratitle}[1]{\vspace{.05in} \noindent\textbf{#1}}

\providecommand{\ie}{\emph{i.e.,} }
\providecommand{\eg}{\emph{e.g.,} }

\newcommand{\boldred}[1]{\textcolor{red}{\textbf{#1}}}

\newcommand{\haroondraft}[1]{\textcolor{purple}{#1}}

\newcommand{\kostasdraft}[1]{\textcolor{cyan}{\textbf{#1}}}

\usepackage{bbm}

\usepackage[flushleft]{threeparttable}
\setlist[enumerate]{wide=\parindent}

\def\ToolX{\textsc{Harpo}\xspace}

\begin{document}

%
\title{\ToolX: \\ Learning to Subvert Online Behavioral Advertising}

\author{\IEEEauthorblockN{Jiang Zhang, Konstantinos Psounis}
\IEEEauthorblockA{University of Southern California\\
\{jiangzha, kpsounis\}@usc.edu}
\and
\IEEEauthorblockN{Muhammad Haroon, Zubair Shafiq}
\IEEEauthorblockA{University of California, Davis\\
\{mharoon, zubair\}@ucdavis.edu}}


%


\IEEEoverridecommandlockouts
\makeatletter\def\@IEEEpubidpullup{6.5\baselineskip}\makeatother
\IEEEpubid{\parbox{\columnwidth}{
    Network and Distributed Systems Security (NDSS) Symposium 2022\\
    27 February - 3 March 2022, San Diego, CA, USA\\
    ISBN 1-891562-74-6\\
    https://dx.doi.org/10.14722/ndss.2022.23062\\
    www.ndss-symposium.org
}
\hspace{\columnsep}\makebox[\columnwidth]{}}

\maketitle

\begin{abstract}
Online behavioral advertising, and the associated tracking paraphernalia, poses a real privacy threat.
Unfortunately, existing privacy-enhancing tools are not always effective against online advertising and tracking. 
We propose \ToolX, a principled learning-based approach to subvert online behavioral advertising through obfuscation. 
\ToolX uses reinforcement learning to adaptively interleave real page visits  with fake pages to distort a tracker's view of a user's browsing profile.
We evaluate \ToolX against real-world user profiling and ad targeting models used for online behavioral advertising. 
The results show that \ToolX improves privacy by triggering more than 40\% incorrect interest segments and 6$\times$ higher bid values. 
\ToolX outperforms existing obfuscation tools by as much as 16$\times$ for the same overhead.
\ToolX is also able to achieve better stealthiness to adversarial detection than existing obfuscation tools.
\ToolX meaningfully advances the state-of-the-art in leveraging obfuscation to subvert online behavioral advertising. 
\end{abstract}

\section{Introduction}
\label{sec:introduction}
Online behavioral advertising poses a real privacy threat due to its reliance on sophisticated and opaque tracking techniques for user profiling and subsequent ad targeting  \cite{acar2014web,iqbal2020fingerprinting,papadopoulos2019cookie,Englehardt161MillionSite,englehardt2015cookies,Olejnik2014SellingOff}.
The tracking information compiled by data brokers for the sake of online behavioral advertising is often outright creepy and scarily detailed \cite{federal2014databroker,kim2018jcr,dehling2019consumer,ur2012smart}.
Furthermore, the surveillance capitalism business model of the ``free'' web naturally aligns with mass surveillance efforts by governments  \cite{zuboff2019age,englehardt2015cookies,URL_GOOGLEPRECOOKIES,WAPO2013MobileInterceptTraffic,URL_RFC7258}. 
Finally, beyond privacy, the targeting capabilities of online behavioral advertising are routinely abused for discrimination \cite{NYTimes2018BiasWomen, NewScience2019FBOutsGayMen,Propublica2016FBRace} and manipulation \cite{Matz2019Pnas,medium2018Ethics,CNN2019FBIAds,Intercept2016DeRadicalize}.

To address the privacy concerns of online behavioral advertising, some platforms now allow users to opt in/out of tracking. 
Notably, iOS 14.5 introduced a new App Tracking Transparency feature that requires apps to get permission from users to track them for targeted advertising \cite{APPLE_ATT}.
Unfortunately, the vast majority of data brokers do not give users any meaningful choice about tracking.
The privacy community has also developed privacy-enhancing tools to enable users to outright block online advertising and tracking.
These blocking tools, available as browser extensions such as uBlock Origin \cite{URL_UBLOCKORIGIN} and Ghostery \cite{URL_GHOSTERY}, are now used by millions of users. 
However, advertisers and trackers can often circumvent these blocking tools, \eg by evading blocking rules \cite{iqbal2018adgraph,wang2016webranz,merzdovnik2017block,alrizah2019errors,chendetecting,vastel2018filters,le21cvinspector} or bypassing these tools altogether \cite{bashir2018tracking,daocharacterizing,subramani2020measuring}.
Thus, blocking is not the silver bullet against online behavioral advertising.

The privacy community has recently started to leverage obfuscation to subvert online behavioral advertising without resorting to outright blocking or to complement blocking \cite{howe2017engineering,URL_TRACKTHIS}.
Unfortunately, existing privacy-enhancing obfuscation tools have limited effectiveness.
For example, AdNauseam  \cite{URL_ADNAUSEAM,howe2017engineering} by Howe and Nissenbaum  obfuscates a user's browsing profile by randomly clicking on ads.
As another example, TrackThis \cite{URL_TRACKTHIS} by Mozilla obfuscates a user's browsing profile by visiting a curated set of URLs.
The effectiveness of these (and other relevant approaches, e.g., \cite{Degeling18bluekaiwpes,biega2017privacy,beigi2018protectinguserprivacy}, discussed in \S \ref{sec: related}) is limited because they are not principled and also prone to adversarial detection.



We propose \ToolX, a privacy-enhancing system that helps users obfuscate their browsing profiles to subvert online behavioral advertising. 
To this end, \ToolX interleaves real page visits in a user's browsing profile with fake pages. 
Unlike prior obfuscation tools, \ToolX takes a principled learning-based approach for effective obfuscation. 
More specifically, \ToolX leverages black-box feedback from user profiling and ad targeting models to optimize its obfuscation strategy.
In addition, and equally importantly, \ToolX's obfuscation is able to adapt to the user's persona.
This principled and adaptive approach helps \ToolX minimize its overhead by introducing the most effective fake page visit at each opportunity, and enhances its stealthiness against adversarial detection.

At its core, \ToolX leverages reinforcement learning (RL) to obfuscate a user's browsing profile. 
\ToolX trains an RL-based obfuscation agent by analyzing a user's browsing profile using an embedding and then optimizing the reward by interacting with a black-box user profiling or ad targeting model. 
\ToolX's trained RL agent is then used to introduce fake page visits into the user's browsing profile at a budgeted rate.
A key challenge in designing \ToolX is that the state space of the underlying Markov Decision Process (MDP) is prohibitively large.
We use a recurrent neural network (NN), together with a convolutional NN as an encoder, and two fully connected NNs as decoders to alleviate the state space explosion of the MDP.
Another key challenge in implementing \ToolX is that we have limited black-box access to real-world user profiling and ad targeting models.
We overcome this challenge by training surrogate user profiling and ad targeting models and leveraging them to train the RL agent.

We evaluate \ToolX against real-world user profiling and ad targeting models \cite{URL_ORACLE,Pachilakis2019headerbiddingimc}.
We find that \ToolX is able to successfully mislead user profiling models by triggering more than 40\% incorrect interest segments among the obfuscated personas. 
We find that \ToolX is able to successfully mislead ad targeting models by triggering 6$\times$ higher bid values.
We also find that \ToolX outperforms existing obfuscation tools by as much as 16$\times$ for the same overhead and by up to 13$\times$ with 2$\times$ less overhead.
We also demonstrate that \ToolX achieves better stealthiness against adversarial detection than existing obfuscation tools.


We summarize our key contributions as follows:

\begin{itemize}

\item We propose \ToolX, a principled RL-based approach to adaptively obfuscate a user's browsing profile.

\item We develop surrogate ML models to train \ToolX's RL agent with limited or no black-box access to real-world user profiling and ad targeting models.

\item We demonstrate the success of \ToolX against real-world user profiling and ad targeting models in terms of privacy, overhead, and stealthiness.

\end{itemize}

\noindent \textbf{Paper Organization:} The rest of the paper is organized as follows.
Section \ref{sec: threat model} describes the threat model.
Section \ref{sec: approach}  presents the design and implementation of \ToolX.
We describe the experimental setup, including data collection and training process for \ToolX and baselines, in Section \ref{sec: setup}. 
Section \ref{sec: evaluation} presents the evaluation results.
We discuss ethical issues and limitations in Section \ref{sec:ethics}.
Section \ref{sec: related} summarizes prior literature before concluding with Section \ref{sec:conclusion}.

\section{Threat Model}
\label{sec: threat model}
The goal of the obfuscation system is to protect the privacy of a \textit{user} against profiling and targeting models of a \textit{tracker}.
To this end, the obfuscation system interleaves the user's real page visits with fake pages to distort the tracker's view of the user's browsing profile.

\paratitle{User.}
The user's goal is to routinely browse the web while misleading the tracker so it cannot accurately profile their interests and subsequently target ads.
Users protect themselves by installing a modified user agent (\ie browser or browser extension) that obfuscates a user's browsing profile by inserting fake page visits into the user's real page visits.
The design goals for this obfuscation system are:

$\bullet$ it should be \textbf{seamless} in that it should not require any modifications to the user's real page visits.

$\bullet$ it should be \textbf{principled} in that misleading the tracker's profiling and targeting models is guaranteed. 

$\bullet$ it should be \textbf{adaptive} to the real page visits so the fake page visits are not trivially always the same.

$\bullet$ it should be \textbf{stealthy} so that it is not possible for the tracker to detect obfuscation and discount fake page visits.

$\bullet$ it should have \textbf{low overhead} to preserve user experience. 

We start by assuming that the user has black-box access to the actual profiling and targeting models of the tracker. 
To relax this assumption, we assume that the user can train a \textit{surrogate model} that is different from the actual model but can reasonably replicate its output.

\paratitle{Tracker.} 
The tracker is typically a third-party that is included by first-party publishers to provide advertising and tracking services on their sites.
We assume that the tracker is able to link the user's different page visits by using well-known cross-site tracking techniques such as cookies or browser fingerprinting.  
We consider a strong threat model by assuming that the tracker has complete coverage of a user's browsing profile.
While a tracker typically does not have complete coverage, prior literature has shown that some trackers indeed have significant coverage of top sites and that even trackers with smaller individual coverage collaborate with each other to improve their coverage \cite{nikiforakis2013cookieless,englehardt2015cookies,yu2016tracking,papadopoulos2019cookie}. 
The tracker is also assumed to have substantial computational resources to train machine learning models on the user's browsing profile to effectively profile the user's interests and target relevant ads \cite{URL_CRITEOAI}.\footnote{Note that while tracking may take place via additional modalities like location, browsing profile based profiling and targeting remains one of the most important modalities currently used in the online advertising ecosystem. See \S \ref{subsec:limitation} for more discussion.}

We also assume that the tracker's goal is to train machine learning models to profile and target arbitrary users rather than a particular user with a known identity (e.g., email address, account identifier).
In the latter case, the tracker can trivially gather information by collaborating with a first-party publisher (e.g., social network or e-commerce site). 
We assume that it is not the case.
Even when this assumption is invalid, we contend that a privacy-conscious user would be able to leverage data deletion requests under privacy regulations, such as GDPR \cite{URL_GDPR} or CCPA \cite{URL_CCPA}, to remove their identity or information.
In summary, we assume that the tracker does not have the user's non-obfuscated browsing profile to begin with.

\section{Proposed Approach}
\label{sec: approach}
In this section, we present the design and implementation of our proposed obfuscation approach called \ToolX.

\subsection{Overview}
\ToolX inserts a fake page visit at random times, where the percentage of fake versus real user's page visits is a configurable system parameter. 
We refer to the corresponding URLs as \emph{obfuscation} versus \emph{user} URLs, respectively. 
Every time a fake page is to be visited, \ToolX needs to decide which URL to pick as the obfuscation URL. 
The decision of \ToolX depends on the user's current browsing profile, which is modeled as a random process because neither the user URLs nor the sequence of obfuscation and user URLs are deterministic. 
Clearly, \ToolX's decisions impact the accuracy of the tracker's profiling and targeting models---the less their accuracy the better the effectiveness of \ToolX.

We formulate the selection of obfuscation URLs as a Markov Decision Process (MDP) which selects URLs to maximize the distortion of the tracker's estimate of the user's interests.
This MDP is not analytically tractable because the exact mechanism that trackers use to create profiles is unknown. 
Moreover, even if a good tracker model were available, the state space of this MDP is prohibitively large to be solved analytically. 
The obvious choice hence is to use Reinforcement Learning (RL) \cite{rl}, which uses feedback from the tracker to  train the RL agent that then selects suitable obfuscation URLs.

\begin{figure}[!t]
    \centering
    \includegraphics[width=\linewidth]{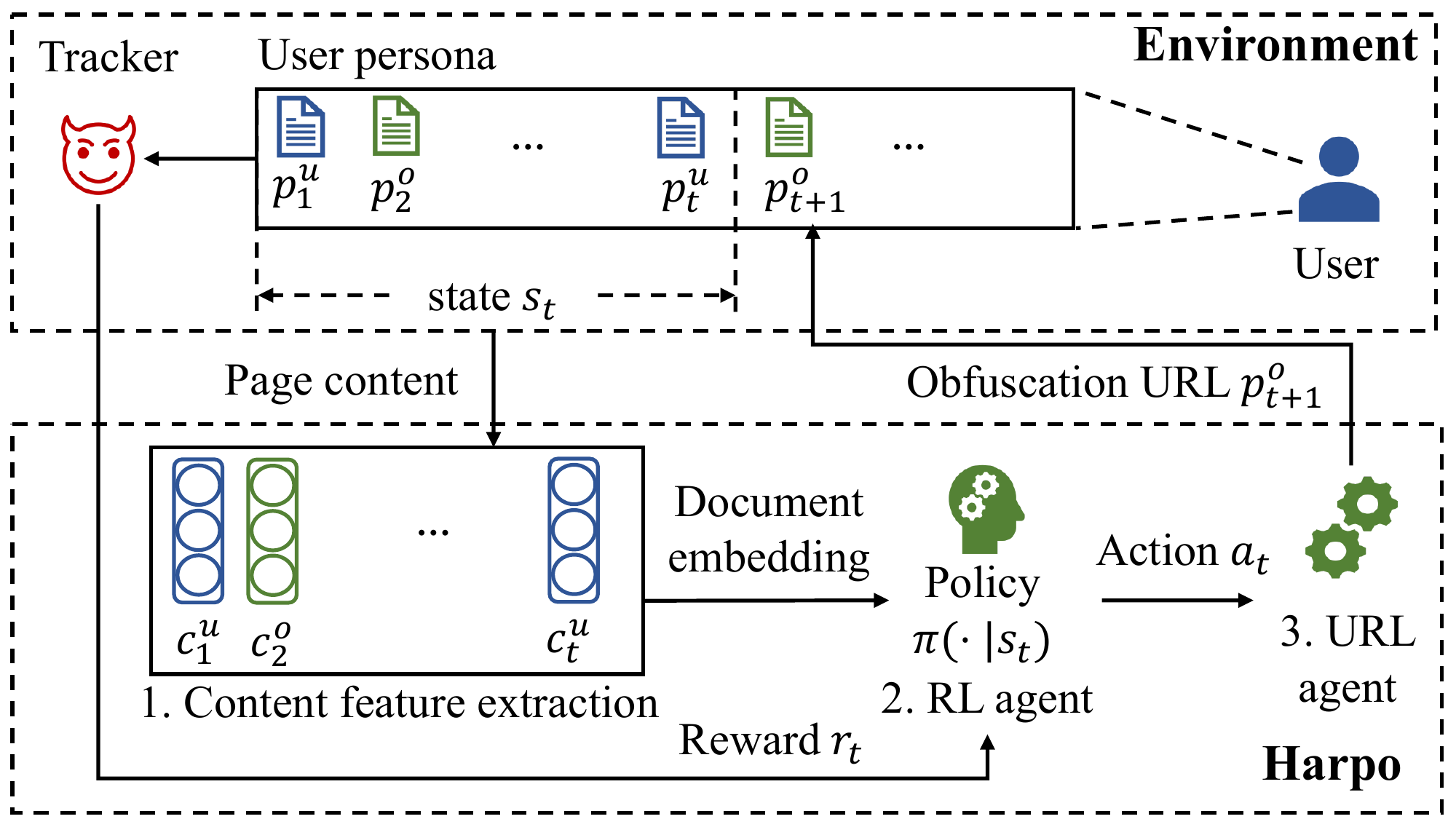}
    \caption{Overview of \ToolX's workflow. Note that the $i$th URL, $p_i$, can be a user or an obfuscation URL, denoted by $p_i^u$ and $p_i^o$ respectively. $c_i^u$ / $c_i^o$ represents the  embedding vector of the $i$th real (blue) / fake (green) page, respectively. 
    }
    \label{fig:overall}
    \vspace{-.2in}
\end{figure}

Figure \ref{fig:overall} illustrates \ToolX's workflow. 
\ToolX starts by parsing the content from the pages of the visited URLs, and featurizes it using an embedding. 
It then trains an RL agent that selects obfuscation URLs to optimize its reward based on the extracted  features. 
After training, the RL agent is used by a URL agent that inserts obfuscation URLs, interleaving them with user URLs.

\subsection{System Preliminaries}

\paratitle{User persona.} 
%
We define user persona simply as the set of visited URLs. 
We denote the user URL set, obfuscation URL set, and the full URL set by $\mathcal{P}^u$, $\mathcal{P}^o$, and $\mathcal{P}$ respectively, where $\mathcal{P}=\mathcal{P}^u\cup\mathcal{P}^o$. 
Since we are interested in the URL selection rather than the time interval between consecutive URLs, we focus at URL insertion times and work with time steps.
We represent a user persona at time step
$t$ by $P_t=[p_1, \cdots, p_t]$, where $p_i$ represents the $i$th visited URL.
At every time step $i$, $1\leq i\leq t$, we select
an obfuscation URL with probability $\alpha$, denoted by $p_i^o\in\mathcal{P}^o$, or
a user URL with probability $1- \alpha$, denoted
by $p_i^u\in\mathcal{P}^u$, where $\alpha$ is a parameter to control the percentage  
of obfuscation URLs.
%
For each obfuscated persona there is a corresponding base persona without the obfuscation URLs, and we denote those personas by $P_t^o$ and $P_t^u$ respectively. 
%


\paratitle{User profiling and ad targeting models.}
Many advertisers provide \textit{interest segments}, such as ``travel-europe", ``pets-dogs", ``health-dementia", inferred by their user profiling models for transparency \cite{Bashir19adpreferencemanagersndss,urban2019study}.
Furthermore, the bids placed by advertisers as part of the real time bidding (RTB) protocol are often visible in the browser \cite{papadopoulos2017if}.
We leverage this black-box access to user profiling (\ie interest segments) and ad targeting models (\ie bid values\footnote{The bid values are measured in cost per thousand impressions (also known as CPM).}) to collect the data needed to train our own surrogate models.
Specifically, we extract the interest segments made available by the Oracle Data Cloud Registry \cite{URL_ORACLE} which gathers data based primarily on cookies, and the bid values placed by advertisers
in the {Prebid.js} header bidding implementation  \cite{Pachilakis2019headerbiddingimc,cook20headerbiddingpets}.
Note that Oracle is a well-established data broker that combines data from more than 70 partners/trackers \cite{URL_ORACLE_WHITEBOOK} and is accessible without needing a cumbersome sign-up process. 
Furthermore, segments are added and removed regularly to reflect the user's latest profile.
To collect bids from multiple bidders efficiently, we select the three popular header bidding enabled  sites (\url{www.speedtest.net}, \url{www.kompas.com}, and \url{www.cnn.com}), each of which contains multiple bidders.

\paratitle{Privacy Metrics.} 
\label{subsubsec:loss}
At a high level, we use as privacy metric the distortion in the tracker's user profiling or ad targeting model estimate, expressed as an accuracy loss. 


For the user profiling model, we consider as distortion the addition of new interest segments and the removal of existing interest segments, when comparing the user profile under the base persona and its obfuscated version. 
To define loss metrics along these lines, we consider $N_s$ interest segments in total and define the interest segment vectors of $P_t^o$ and $P_t^u$ as $X^o = [x_{1}^o, …, x_{N_s}^o]$ and $X^u = [x_{1}^u, …, x_{N_s}^u]$ respectively, where $x_{i}^j$, $i\in \{1,\ldots, N_s\}$, $j\in \{o,u\}$, are binary variables representing whether the $i$th interest segment is triggered or not (1: triggered, 0: not triggered). 
Then, we define the following loss for the tracker, which represents the percentage of segments of the obfuscated persona which were not segments of the base persona:
\vspace{-.1in}
\begin{align}
        L_1(X^o, X^u) = \frac{\sum_{i=1}^{N_s} 1_{\{x_i^o=1,x_i^u=0}\}}{\sum_{i=1}^{N_s} x_i^o}.
        \label{eq_L_2}
\vspace{-.1in}
\end{align}
\vspace{-.1in}

Here, the numerator is the number of incorrect segments ($1_A=1$ if $A$ is true and 0 otherwise) and the denominator is the total number of segments of the obfuscated persona.\footnote{Note that $L_1$ is undefined if the obfuscated persona has no segments, which is a corner case of no practical relevance.} 
%


%
We also define a second loss metric for the tracker which aims to quantify the profile distortion: 
\vspace{-.05in}
\begin{align}
        L_2(X^o, X^u) = \sum_{i=1}^{N_s} x_i^o \oplus x_i^u.
        \label{eq_L_1}
\vspace{-.1in}
\end{align}
\vspace{-.15in}

This loss metric equals the number of different segments between the original ($X^o$) and obfuscated ($X^u$) profiles. 
It is maximized if all the interest segments of the base persona are removed by the profile and all the remaining segments are triggered, thus maximally distorting the profile.

For both $L_1$ and $L_2$, the more base persona segments are removed and the more obfuscated persona segments are added, the higher their value. The difference is that, $L_1$ measures the portion of the triggered segments for the obfuscated persona that have no value for the tracker and equals 100\% when all base persona segments are removed, while $L_2$ reports the total number of different segments and thus represents the profile distortion. 
Clearly, the higher the $L_1$ and $L_2$ values are, the lesser the sensitive information contained in the obfuscation profile and the higher the resulting privacy.
For the ad targeting model, we consider as distortion the deviation of the bid values placed by a bidder under the base persona and its obfuscated version. 
It is worth noting that bid values represent a bidder's confidence about whether a user's interests match the targeted ad. 
By manipulating a user's profile to distort bid values, our goal is to make bidders place bids that are inconsistent with the user's interests (e.g., place a high bid value for an ad that is actually irrelevant to the user). 
This inconsistency means that the bidder has incorrectly profiled the user’s interests and thus represents a better privacy outcome for the user.
To maximize the deviation one may attempt to either increase or decrease the bidding values by appropriately selecting obfuscation URLs. 
We choose to attempt to increase the bid values because bidders tend to significantly place more low bid values than high bid values, thus there is much more room to distort low bid values.\footnote{We discuss ethical considerations regarding the potential infliction of economic pain to bidders by this in \S \ref{sec:ethics}.}
%
%

To define practical loss metrics along these lines, we first group the bid values into two classes.
Suppose the mean and variance of all the bid values we collect from a bidder are $\mu$ and $\sigma$ respectively. 
Then, we use $\mu + \sigma$ to split the bid values into low and high value bid classes. 
If a bid value is larger than $\mu + \sigma$, we classify it as high, else we classify it as low.
Now, consider a total of $N_b$ bidders bidding for ads based on the current user browsing profile.
%
%
We define $v_i^j$, $i \in \{1,\ldots,N_b\}$, $j\in\{o,u\}$, as the bid value placed by bidder $i$ for an obfuscated ($j=o$) or non-obfuscated ($j=u$) persona. 
We also define $b_i^j=\mathbbm{1}_{v_i^j\geq\mu_i+\sigma_i}$ to
indicate whether the bid value for bidder $i$ is below ($b_i=0$) or above ($b_i=1$) the threshold $\mu_i+\sigma_i$, where $\mu_i$ and $\sigma_i$ are the mean and variance of bid values placed by bidder $i$.
Then, we use as loss the increase of the proportion of high bids in the obfuscated persona as compared to the corresponding base persona, \ie 
\vspace{-.05in}
\begin{equation}
    L_3(\{b^o_i,b^u_i\}_{i=1}^{i=N_b}) = \frac{1}{N_b}\sum_{i=1}^{N_b} (b_{i}^o-b_{i}^u).
    \label{eq_L_3}
\vspace{-.1in}
\end{equation}

To directly quantify how much the bid values change, we also use the average ratio of bid values of an obfuscated persona over its corresponding base persona and denote it by $L_4$. Specifically,
\vspace{-.05in}
\begin{equation}
    L_4(\{v^o_i,v^u_i\}_{i=1}^{i=N_b}) = \frac{\sum_{i=1}^{N_b}v_{i}^o}{\sum_{i=1}^{N_b}v_{i}^u}.
\end{equation}
\vspace{-.1in}

\subsection{System Model}
\label{subsec:sysmodel}
As discussed earlier, we formulate the selection of obfuscation URLs as an MDP.
In a nutshell, MDP is a standard framework for modeling decision making when outcomes are partly stochastic and partly under the control of a decision maker.
We describe in detail all components of the MDP below.


\paratitle{Obfuscation step.} 
MDPs are discrete-time processes evolving in time steps. 
Recall that we assume time evolves in steps every time a URL is visited.
We refer to a time step as an obfuscation step, if the visited URL at this time step is an obfuscation URL, and use obfuscation steps as the time steps of the MDP.
In the rest of the section we use $t$ to denote obfuscation steps, and let $N_t$ denote the total number of URLs (\ie user and obfuscation URLs) visited by the persona under consideration up to obfuscation time step $t$.
\footnote{Note that to keep the notation in Figure \ref{fig:overall} simple, we have used $t$ to represent time steps corresponding to both user and obfuscation URLs, in a slight abuse of notation.}

\paratitle{State.} 
MDPs transition between states. 
We define the state at obfuscation step $t$ as $s_t=[p_1,\cdots, p_{N_t}]\in\mathcal{S}$, which consists of the visited URLs up to time step $t$, where $\mathcal{S}$ denotes the state space of the MDP. 
Note that this state definition means the state space will grow indefinitely.
Yet we do so because the retention time of URLs by data brokers, including the Oracle Data Cloud Registry, are often in the order of 12 to 18 months \cite{urban2019study}.
Thus, we want to select an obfuscation URL based on the entire browsing profile of a persona. 
While such a state space complicates analytical treatment, as we discuss later in \S \ref{subsubsec:rl}, we use a recurrent model as part of our RL model which allows us to handle this effectively.


\paratitle{Action.} 
MDPs choose an action at each step, based on a policy. 
At obfuscation step $t$, the action $a_t$ is the selection of an obfuscation URL $p_{N_t+1}^o$ from the set of $\mathcal{P}^o$, which is the action space.

\paratitle{State transition.} 
The transition between states of an MDP is dictated by a state transition function $\mathcal{T}(\cdot|\mathcal{S}, \mathcal{A}): \mathcal{S} \times \mathcal{A} \times \mathcal{S} \rightarrow \mathcal{R}$, which outputs the probability distribution of state $s_{t+1}$ given the previous state $s_t$ and the action $a_t$.
Note that state $s_{t+1}$ consists of all visited URLs up to step $t$ ($s_t$), of the obfuscation URL $p_{N_t+1}^o$, and of the user URLs visited between obfuscation step $t$ and $t+1$.

\paratitle{Reward.} 
Every time there is an action, there is a reward associated with it. 
We use as reward of obfuscation step $t$ the difference of the loss of the tracker between this and the previous step.  
Specifically, let $L_t$ denote 
the loss of the tracker at obfuscation step $t$.
$L_t$ can be any of the privacy metrics defined above. 
Then, the reward $r_t$ equals $L_t - L_{t-1}$.

To avoid selecting a small set of high-reward obfuscation URLs repeatedly as this may affect stealthiness, we may use the following reward function which penalizes the selection of the same URLs:
$r_t = L_t - L_{t-1} - \delta*(N(p)-1)$,
where $N(p)$ represents the number of times the obfuscation URL $p$ has been selected in the past and $\delta$ is a parameter controlling the diversity of selected URLs, see \S \ref{subsec:stealth} for related performance results.


\paratitle{Policy.} 
We define the policy of the MDP as $\mathcal{\pi}(\cdot|\mathcal{S}): \mathcal{S} \times \mathcal{A} \rightarrow \mathcal{R}$, where $a_t\sim\pi(\cdot|s_t)$. 
That is, the policy is the probability distribution of the obfuscation URL selection for each state $s_t$. 
Specifically, let $N_o=|\mathcal{P}^o|$ be the total number of available obfuscation URLs.
Then, $\pi(\cdot|s_t)$ is a multinomial distribution with parameter $ A_t = [a_t^1,\cdots,a_t^{N_o}]$, where $a_t^i$ is the probability of selecting the $i$th obfuscation URL, and $\sum_{i=1}^{N_o}a_{t}^i=1$.
We design the policy $\pi(\cdot|s_t)$ with the objective of maximizing the accumulated expected reward over a finite time horizon.
In the RL implementation of the MDP, the finite time horizon equals the number of obfuscation steps during training of the RL agent.

\begin{figure*}[!t]
    \centering
    \includegraphics[width=\linewidth]{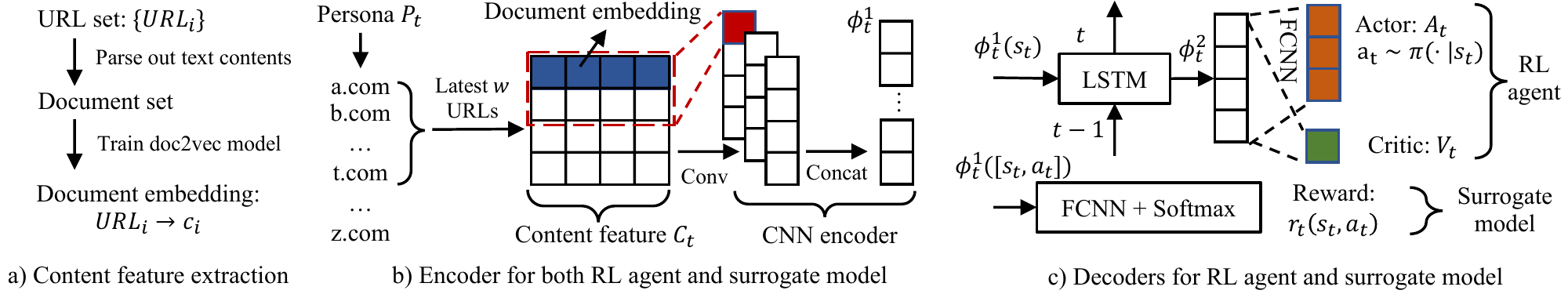}
    \caption{Neural network structures for RL agent and surrogate model. Both the RL agent and the surrogate model take the content features $C_t$ (document embedding output by doc2vec model) of the latest $w$ URLs in user persona $P_t$ as input, and then utilize CNN as encoder to convert $C_t$ into feature vector $\phi_t^i$ as the input of decoders. The decoder of the RL agent is a LSTM followed by two FCNN, representing actor and critic networks respectively. And the decoder of the surrogate model is a FCNN with Softmax activation function, which outputs the binary classification result as the reward for the RL agent.}
    \label{fig:arch}
    \vspace{-.2in}
\end{figure*}

\subsection{System Design}
\label{subsec:design}
\ToolX consists of 4 modules:
(i) a content feature extraction module that converts text to a document embedding, 
(ii) an RL agent which gets document embeddings as input and outputs obfuscation URLs, 
(iii) a surrogate model, trained to replicate real-world user profiling and ad targeting models, which is used for fast training of the RL agent, and 
(iv) a URL agent which inserts obfuscation URLs in the web browsing profile of personas. 
We describe in detail each module below, and present 
an overview of \ToolX's workflow in \F \ref{fig:overall}.

\paratitle{Feature Extraction.}
\label{subsubsec:preprocess}
To extract the features of a visited URL, we train a document embedding model for \ToolX, whose input is the textual content on the page of each visited URL $p_i$ and the output is the document embedding $c_i\in\mathcal{R}^d$ (a real vector with dimension $d$). 
More specifically, as demonstrated in \F\ref{fig:overall}, for each URL in our URL set, we first parse the text content from its rendered HTML page as a text document. 
We then train a doc2vec embedding model \cite{le2014distributed} via unsupervised learning by utilizing the extracted text documents of all URLs in $\mathcal{P}$.
Finally, \ToolX uses the trained doc2vec model to map each URL to an embedding, which represents the  features of the page corresponding to the URL.

We only consider textual content during feature extraction for two reasons.
First, text content is the basis of {HTML} files and can convey the information in the web pages that is relevant for user profiling and ad targeting models \cite{patent_ad_targeting,patent_ad_targeting_2}. 
Second, it is easier and faster to process text content than other types of multimedia content. 
Moreover, since a page typically contains thousands of word tokens, we choose to train a document embedding model instead of word embedding or sentence embedding models, so that the dimension of embedding vectors can be reduced.

\paratitle{RL Structure and Implementation.}
\label{subsubsec:rl}
At a high level, the RL agent consists of a CNN (Convolutional Neural Network) as an encoder, followed by an LSTM (Long-short Term Memory) neural network and two FCNNs (Fully-Connected Neural Networks) as decoder, which represent actor and critic networks respectively. 
The actor network will determine which obfuscation URL to select at each obfuscation step based on the current state, while the critic network will estimate the future cumulative reward based on the current state and the action chosen by the actor network.

Specifically, as illustrated in \Fs \ref{fig:arch}b and \ref{fig:arch}c, the input of the CNN $C_t$ consists of the document embeddings of the latest $w$ URLs ($C_t\in\mathcal{R}^{w\times d}$) and the output of the CNN $\phi_t^1$ is an encoded real vector with $m$ elements ($\phi_t^1\in\mathcal{R}^m$).
$\phi_t^1$ is the input of the LSTM, which outputs a decoded real vector $\phi_t^2$ with $n$ elements ($\phi_t^2\in\mathcal{R}^n)$.
$\phi_t^2$ will further be the input of the actor and critic networks, which output the probability distribution of selecting each obfuscation URL $A_t\in\mathcal{R}^{N_o}$ (recall there are $N_o$ obfuscation URLs in total) and the estimate of the expectation of the future accumulated reward $V_t\in\mathcal{R}$ (a real number), respectively. 
We train the actor critic networks via the A2C (Advantage Actor and Critic) algorithm \cite{openai2017a2c}, which is one of the most popular on-policy RL algorithms. Note that we select on-policy RL algorithms since they are more memory efficient and adaptive to the dynamic web environment as compared to off-policy RL algorithms.

We choose CNN as the encoder of the document embedding since it has fewer training parameters compared with other Deep Neural Networks (DNNs) and prior works demonstrate its effectiveness on text classification (\eg \cite{kim2014convolutional}). 
Furthermore, we use an LSTM because it is a recurrent neural network which allows us to maintain information about the whole browsing profile despite the input to the RL agent being the $w$ most recent pages only.
Prior work has also used an LSTM when the MDP state is only partially observable by an RL agent
\cite{hausknecht2015deep,xu2019experience}. 
Note that The RL agent’s input at each obfuscation step is an embedding matrix consisting of a sequence of doc2vec embeddings. Adding a CNN before the LSTM can extract the local features of the embedding matrix efficiently and reduce the input feature space of the LSTM (from a 2D matrix to a vector) at each obfuscation step. Prior research \cite{zhou2015c} has also demonstrated the effectiveness of combining CNN with LSTM.

\paratitle{Surrogate model.} 
\label{subsubsec:surrogate_model}
To train the RL agent, we would need ample access to a real-world user profiling or ad targeting model. 
However, as outlined in the threat model, we may have limited or no access to the real-world user profiling or ad targeting models in practice.
To address this issue, we propose to train surrogate models that can reasonably replicate the output of real-world user profiling or ad targeting models.
These surrogate models are then used to train the RL agent. 
The surrogate models also help improve the efficiency of RL agent training by providing a virtual environment, which is much faster than querying real-world user profiling or ad targeting models.\footnote{While profile registries like the Oracle Data Cloud Registry are required by law to allow users access to their profiles, these profiles may be updated every few days. Thus, it would take months to collect enough samples to train the RL agent solely by accessing such registries.}
Next, we describe in detail the surrogate models for user profiling and ad targeting systems.

For the user profiling model, we train a separate model for each interest segment in the Oracle Data Cloud Registry to predict whether this interest segment will be triggered by the most recent $w$ URLs in the web browsing profile of a persona. Note that we use the latest $w$ URLs rather than the complete web browsing profile, because it is hard to accurately train models with very long and variable length inputs.
More precisely, for a user persona with a  browsing profile of $N_t$ URLs at obfuscation step $t$, $P_{N_t}=[p_1,\cdots,p_{N_t}]$, 
we extract the document embedding of the latest $w$ URLs, $C_t = [c_{N_t-w+1}, …, c_{N_t}]$, and feed them as input into the model. 
The model, which we refer to as a segment predictor henceforth, outputs a 1 if the segment is expected to be triggered, and a 0 otherwise. 
%

For the ad targeting model, as discussed already, we first group continuous bid values into a low- and a high-bid class. 
Then, we train a binary classifier to predict the bid class and refer to this model as the bid predictor. 
Similar to the segment predictor models, the bid predictor takes $C_t$ as the input and outputs either 0 (low bid class) or 1 (high bid class).

The detailed structure of surrogate models are demonstrated in \Fs \ref{fig:arch}b and \ref{fig:arch}c, which consist of a CNN and FCNN with Softmax activation.
Specifically, the CNN has the same structure as that in the RL agent, which takes $C_t$ as input and outputs $\phi_t^1$ (see \S \ref{subsubsec:rl}). 
The decoder, which is the FCNN, takes $\phi_t^1$ as input and outputs the binary classification value (0 or 1) of each surrogate model.

To train the bid and segment predictors, we start by randomly constructing a set of user personas. 
Then, we collect training data (by the Oracle Registry for the user profiling model and multiple bidders for the ad targeting model) and use supervised learning,
see \S \ref{sec: setup} for more details.

\paratitle{URL Agent.} 
\label{subsubsec:plugin}
The URL agent creates user personas consisting of both user and obfuscation URLs through an i.i.d random process. 
At each time slot, with probability $\alpha$ the URL is an obfuscation URL selected by \ToolX and with probability $1-\alpha$ it is a user URL, randomly picked from the user URL set. %
In practice, the URL agent would not generate user and obfuscation URLs in discrete time slots. 
Instead, it would estimate the arrival rate of user URLs, call it $\lambda^u$. 
Then, to target an obfuscation ``budget" $\alpha$, it would create a random process with arrival rate $\lambda^o=\frac{\lambda^u*\alpha}{1-\alpha}$ to specify the insertion times 
of obfuscation URLs. 
For example, a Poisson process with rate $\lambda^o$ can be used for that purpose, or, a non-homogeneous Poisson process can be used to adapt $\lambda^o$ (more precisely, $\lambda^o(t)$ in this case) to the current user behavior (\ie if the user is not engaged in an active browsing session, very few or no obfuscation URLs would be inserted).

\subsection{System Implementation}
We implement \ToolX as a browser extension. 
Its architecture has a passive \textit{monitoring} component and an active \textit{obfuscation} component. 
The monitoring component uses a background script to access the webRequest API to inspect all HTTP requests and responses as well as a content script to parse the DOM and extract innerHTML \cite{URL_WEBEXTENSION}.
This capability allows us to implement the content extraction module, which is responsible for computing document embedding for each visited page. 
The monitoring component sends this information to the obfuscation component, which is responsible to implement the other 3 modules of \ToolX (RL agent, surrogate model, and URL agent). 
The RL agent and surrogate model modules run in the background, the former to select an obfuscation URL that is visited by the URL agent module, and the later to train the RL agent.
To visit the obfuscation URL in the background so user experience is seamless, we open the URL in a background tab that is hidden from the user's view.\footnote{In \S \ref{subsec:limitation}, we discuss how to prevent a tracker from using side-channels associated with background tabs to detect \ToolX.}
Note that our implementation does not simply use AJAX to simulate clicks \cite{howe2017engineering}, it realistically loads pages by executing JavaScript and rendering the page content. 
\ToolX's browser extension is implemented to minimize adverse impact on user experience.
We evaluate the system overhead of \ToolX's browser extension implementation later in \S\ref{subsec:budget}.

\vspace{-.05in}
\section{Experimental Setup}
\label{sec: setup}

\subsection{User Persona Model}
\label{subsec: persona model}
We need to gather realistic web browsing profiles to experimentally evaluate \ToolX and baselines.
While we could try to directly use real-world web browsing traces, this would pose two problems from a practical standpoint.
First, we need to restrict the total number of distinct URLs to a manageable number that we can crawl in a reasonable amount of time. 
Second, it is preferable to train a model that can work for general user types than individual users. 
To address these problems, we first use real-world web browsing traces to train a user persona model, and then use this model to generate a large number of web browsing profiles from a manageable pool of distinct URLs and user types.

Specifically, we start with the AOL dataset \cite{pass2006picture} which consists of millions of distinct URLs and web browsing profiles of millions of users.\footnote{
While the AOL dataset is somewhat dated, it is one of the largest publicly available datasets of real-world user browsing profiles and captures well the browsing behavior of a large, diverse set of users.}
We then randomly sample users with more than 100 visited URLs each, and leverage
WhoisXMLAPI \cite{URL_WhoisXMLAPI} to map each URL into one of the 16 IAB categories from Alexa \cite{URL_ALEXASITESBYCATEGORY}.
We observe that real web browsing profiles consist of URLs from a handful of preferred URL categories.
Motivated by this, we use a Markov Chain (MC) model to generate web browsing profiles as follows: a MC state dictates the category from which a URL is selected. 
We assign a separate state to each of the most popular categories, and a single state collectively to the rest of the categories. 
As the MC transits from state to state, a URL is randomly selected from the URL category (or categories) that corresponds to the current state.

\begin{figure}
\begin{subfigure}{.23\textwidth}
    \includegraphics[width=.99\linewidth]{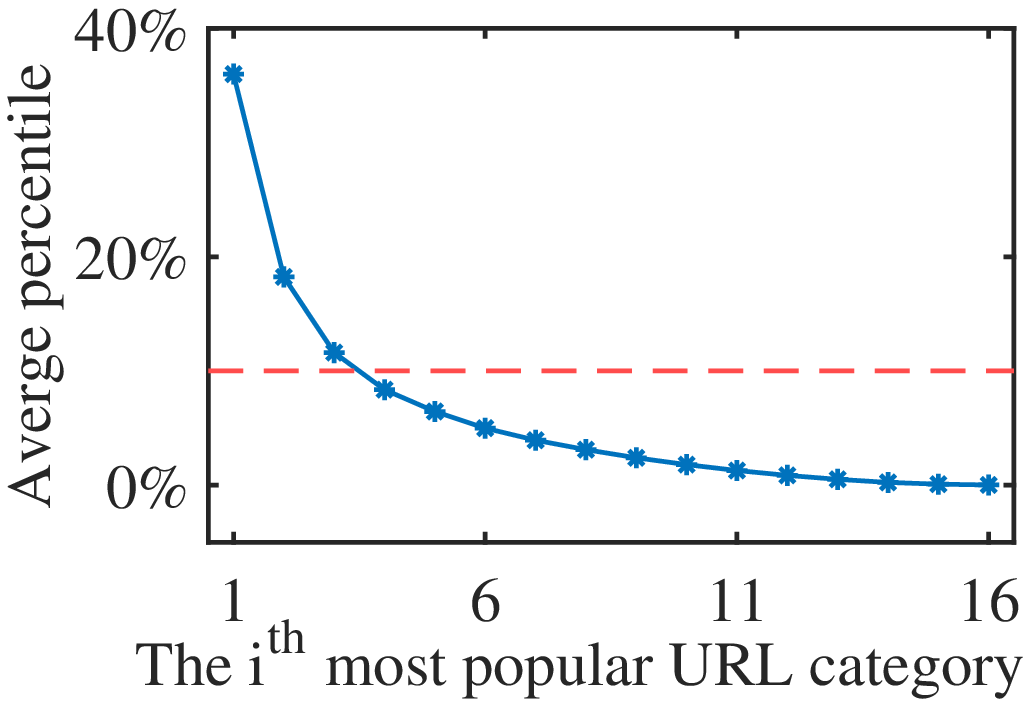}
    \caption{Percentage of URLs from $i^{th}$ most popular category.}
    \label{fig:mc_state}
\end{subfigure}
\begin{subfigure}{.23\textwidth}
    \includegraphics[width=.99\linewidth]{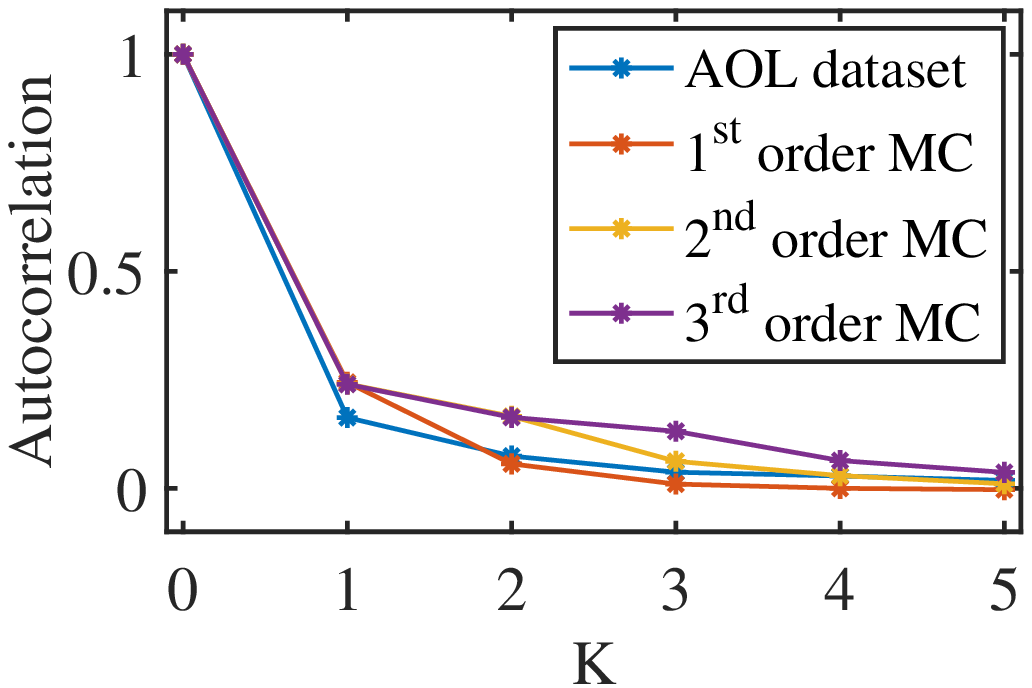}
    \caption{Autocorrelation for different order MCs.}
    \label{fig:mc_auto}
\end{subfigure}
\vspace{-.1in}
\caption{Selecting parameters of the MC model. Note that the autocorrelation with lag $K$ measures the correlation between states that are $K$ time steps apart.}
\label{fig:mc_analysis}
\vspace{-.25in}
\end{figure}

To specify the model parameters, first we need to decide how many popular categories will have their own state.
We do so by assigning a separate state to categories whose URLs represent more than 10\% of the total URLs in the dataset. 
Figure \ref{fig:mc_state}
plots the percentage of URLs in a user's web browsing profile from the $i^{th}$ most popular URL category for this user, averaged over all users. From the figure we conclude that the 3 most popular categories satisfy our criteria. Thus, we set the total number of states of the MC to 4, one for each of the 3 most popular categories and one collectively for the 13 remaining categories.

\begin{figure}
    \centering
    \includegraphics[width=.95\linewidth]{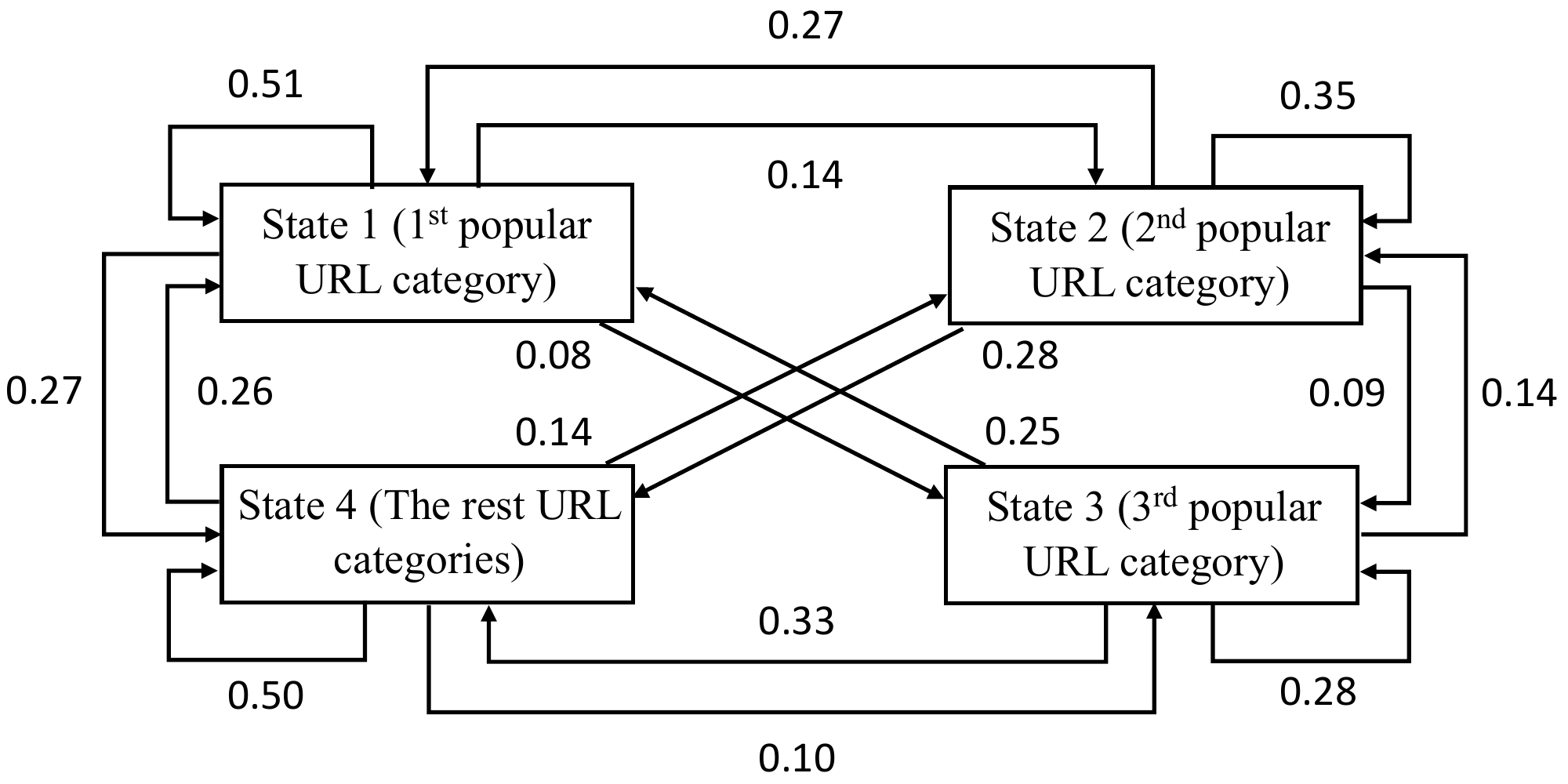}
    \vspace{-.05in}
    \caption{The MC model and its state transition probability diagram for simulating user personas.}
    \label{fig:MC_model}
\vspace{-.25in}    
\end{figure}

Next, we need to decide the order of the MC. In general, a higher order MC has the ability to model longer term correlations between states, as the transition probability from one state to another for a $j$th-order MC depends on the $j$ most recent states. That said, the higher the order the higher the complexity of the MC, as the state space grows exponentially, see, for example,   \cite{gagniuc2017markov}.
Following standard practice, we use the autocorrelation function to measure the correlation in the AOL dataset and experiment with different order MCs to identify the smallest order required for a good fit. Figure \ref{fig:mc_auto} shows that a $1$st order MC is enough to achieve a good fit.  
Last, given the order and number of states of the MC, we fit the stationary distribution and transition probabilities of our MC model to the statistics of the dataset (see \F \ref{fig:MC_model} for the final MC model).

In the rest of the paper, we use the aforementioned model to generate web browsing profiles for user personas.
Since the most popular categories are not necessarily the same for each user persona, we select the 100 most common combinations of the 3 most popular URL categories from the AOL dataset and define 100 user types. Then, every time we want to generate a web browsing profile for a user persona, we randomly select one user type which sets the specific 3 most popular categories for this user, and use the MC model to generate the user URLs as described above.

\subsection{Data Collection and Preparation}
\label{subsec: personalurls}

\paratitle{Persona URLs.}
The web browsing profiles of user personas consist of user and obfuscation users. User URLs are generated by the user persona model described above. Note that for each of the 16 IAB categories, we keep the 100 most popular URLs within each category as ranked by Alexa \cite{URL_ALEXASITESBYCATEGORY}, thus there are a total of 1600 user URLs to pick from every time a user URL is selected.

Obfuscation URL categories depend on the obfuscation scheme (\ToolX or one of the baseline approaches described in \S \ref{subsec: baselines}) and we consider three different categories: TrackThis, AdNauseam, and intent URL categories.
The TrackThis category contains 400 obfuscation URLs from \cite{URL_TRACKTHIS}. 
For the AdNauseam category, we collect all the third-party URLs from the 1,600 pages corresponding to the 1600 URLs of the 16 IAB categories we described above, and identify advertising URLs using EasyList \cite{URL_ADBLOCK}. In total, we collect 2,000 advertising URLs. 
%
For the intent category, we randomly crawl 1,930 product URLs through Google shopping (10 URLs for each one of the 193 shopping categories, which we will refer to as intent URL subcategories henceforth).

\paratitle{Data collection for surrogate models.}
We construct 10,000 personas to collect data from real-world user profiling and ad targeting models in order to train the surrogate models. The proportion of obfuscation URLs, $\alpha$, in each persona varies between 0 and 0.2. 

Collecting data from real-world models is a costly operation. Thus, we determine the suitable length of a persona based on the following analysis, keeping data collection efficiency in our mind.
Let $N$ be the average number of URLs per persona, which we wish to determine. 
Let $n$ be the fraction of personas for which we are able to collect some feedback (\ie the trackers return no feedback for the rest). 
$10,000\cdot n$ is the total number of personas we can collect feedback for and $10,000\cdot N$ is the total number of URLs among all personas. 
We choose to select $N$ such that we maximize $n/N$ for the following reason:
While longer personas with more URLs will likely trigger more feedback, computational overheads (\eg CPU and memory) are also proportional to the total number of URLs.
Thus, the most efficient choice is to maximize the feedback we collect per URL, and $n/N$ represents the number of personas with non-empty feedback per URL. 
The above procedure yields a value of $N$ equal to 20, and we use the MC model described above to select the user URLs, and \ToolX to select obfuscation URLs from the intent URL category, for a total of 20 URLs per user persona. 

Using these personas we collect feedback from real-world user profiling and ad targeting models as follows:
For each persona, we start with a fresh browser profile in OpenWPM \cite{Englehardt161MillionSite}.
For ad targeting, we access bidding sites to collect the triggered bids immediately after visiting the 20 URLs. 
For user profiling, since we observe that it takes on average two days for triggered interest segments to appear in Oracle Data Cloud Registry, we save the browser state after visiting the 20 URLs and reload it after 2 days to collect the triggered interest segments.
In total, we collect 184 different interest segments from the Oracle Data Cloud Registry and bids placed by 10 different bidders on 16 different ad slots. 
Note that for each bidding site, there could be multiple bidders that might place different bids for different ad slots.

\paratitle{Data preparation for surrogate models.}
We first clean the data by removing unrelated interest segments such as those related to geographic location or device type, and by removing zero bids. 
Then, for each user persona for which we collected some feedback, we extract content features from the visited web pages, concatenate the document embedding vectors of all visited URLs of the persona into an embedding matrix, and use this matrix as the input to the surrogate models.
We use a surrogate model for each interest segment, where the label is a binary variable with 1/0 representing that the user persona will/will not trigger the segment, respectively.
We also use a surrogate model for each bidder and ad slot pair, where the label is a binary variable with 1/0 representing that the user may trigger a high/low bid, respectively.

\S \ref{subsec:trainingtesting} discusses how we train the surrogate models using supervised learning. 
Let a dataset refer to all the user persona embedding matrices and the associated labels collected.
If the percentage of labels with value 1 in a dataset is less than 5\%, we remove it because it is likely not sufficient for training surrogate models later. 
We end up with 121 interest segment datasets and 50 bid datasets for training a total of 171 surrogate models.

\paratitle{Data collection for RL agent.}
\label{subsubsec:data-rl}
We construct 15,000 personas, 50 for each of 300 training rounds of the RL agent, to train the RL agent.
Each persona consists of 100 URLs.
Recall that user URLs are selected using the MC model from the IAB categories, and the obfuscation URLs are selected from the intent category based on the actions generated by the RL agent.
The first 20 URLs are selected randomly from the user URL set for initialization. 
The remaining 80 URLs are either obfuscation URLs (with probability $\alpha$) or user URLs (with probability $1-\alpha$).
Thus, we have on average $80\cdot \alpha$ obfuscation URLs
per persona. 

A word on the selection of the $\alpha$ value is in order. 
Clearly, the smaller the $\alpha$ the lower the overhead. Also, one may conjecture that
the smaller the $\alpha$ the higher the stealthiness. 
However, too small of an $\alpha$ value may not yield enough obfuscation URLs to have a large impact.
We start our evaluation by choosing $\alpha=0.1$ for both user profiling and ad targeting.
In \S \ref{subsec:budget} and \S \ref{subsec:stealth} we study the impact of $\alpha$ on obfuscation effectiveness, overhead and stealthiness.

\paratitle{System configuration.} 
We use OpenWPM \cite{Englehardt161MillionSite} to implement our crawling system in an automated and scalable manner.
The experiments are run on servers with 32 CPUs and 128GB memory on an AMD Ryzen Threadripper 3970X server with 3.7GHz clockspeed. 
%

\subsection{Training and Testing}
\label{subsec:trainingtesting}
We report the neural network parameter values of surrogate model and RL agent in \T \ref{tab:param}, and describe the training and testing process of the surrogate models and the RL agent in this subsection.

\paratitle{Surrogate models.} 
For each of the 176 surrogate models (121 interest segment and 55 bid models), we utilize 80\% of the data collected from the 10,000 personas for training and 20\% for testing. 
We train each model via stochastic gradient descent \cite{bottou2010large} with a batch size of 32 personas for 30 training rounds, where all the training data are used once at each training round.

\paratitle{RL agent.}
We train and test the RL agent with the data collected from the 15,000 personas.
Specifically, we train the RL agent using the surrogate models to collect the reward and run the training for 300 rounds.
We test the RL agent using surrogate models for 10 rounds.
At each training or testing round we generate a batch of 50 personas.

In addition to testing the RL agent using surrogate models, we also test it against real-world user profiling and ad targeting models.
To this end, we create 100 personas with 100 URLs each. 
For each persona we start with a fresh browser profile in OpenWPM \cite{Englehardt161MillionSite}.
For ad targeting, we immediately collect the triggered bid values as we visit the 100 URLs of the persona.
For user profiling, we save the browser state after visiting all 100 URLs of the persona, and wait for 2 days to access the Oracle Data Cloud Registry and collect the triggered interest segments.

Recall that \ToolX selects obfuscation URLs from the intent URL category which consists of 1930 URLs (10 URLs from each of the 193 intent URL subcategories). For scalability reasons we wish to reduce the number of possible decisions, thus implement the RL agent to select on of the intent URL subcategories, and then \ToolX randomly selects one of the URLs within the selected intent URL subcategory. Note that by construction, URLs within the same intent URL subcategory have similar content.

\paratitle{Training cost.}
We note that it takes about 2 minutes to train the surrogate model from scratch and less than 1 minute to train the RL agent per-round on our server. While the exact training time would vary depending on the user’s system specifications, we do not expect it to take longer than a few minutes.

\begin{table}[htb]
    \small
    \centering
    \caption{Parameter values of neural networks for RL agent and surrogate model in \ToolX.}
    \begin{tabular}{p{1.8in}p{1.3in}}
    \hline
         Parameter description               &  Configuration \\ \hline
         Dimension of document embedding   & $d=300$ \\
         Dimension of CNN input            & $w\times d=20\times 300$ \\
         Kernel size in CNN    & $\{i\times i\times 1\times 100\}_{i=3,4,5}$ \\
         Dimension of encoder vector       & $m=300$ \\
         Dimension of decoder vector       & $n=256$ \\
         Dimension of actor's output       & $N_o=193$\\
         \hline
    \end{tabular}
    
    \label{tab:param}
\end{table}
\subsection{Accuracy of Surrogate Models}
\label{subsec:surrogate}
We study the accuracy of the surrogate models we trained for user profiling and ad targeting and report the true positive rate (TPR) and false positive rate (FPR) metrics. 
Out of the 121 interest segment models we 
select the 20 most accurate, and out of the 55 bid models we select the 10 most accurate. 
We then use those models to train and evaluate \ToolX in the context of user profiling and ad targeting.



\paratitle{User profiling.} In general, the trained surrogate user profiling models have reasonable accuracy. As reported in \T\ref{tab:predictor}, the average FPR and TPR of the 20 most accurate surrogate user profiling models are 3.92\% and 96.57\% respectively. 
The FPRs of these 20 surrogate user profiling models ranges from 1.82\% to 19.23\% and the TPRs vary from 81.43\% to 100.00\%.
Last, among the 20 datasets training the top 20 surrogate user profiling models, the percentage of data points with label value 1 (positive data, indicating the segment is triggered) varies from 3.91\% to 15.87\%, with an average value of 7.04\%.

\paratitle{Ad targeting.} Compared with user profiling surrogate models, we observe that ad targeting surrogate models are less accurate in general. This is expected since the bids placed by each bidder are likely affected by other auction dynamics, and their values have larger variance making it more difficult to predict accurately \cite{papadopoulos2017if}. 
However, we still obtain 10 surrogate ad targeting models with good accuracy, which achieve 18.28\% FPR and 73.43\% TPR on average as shown in \T\ref{tab:predictor}. 
The FPRs of these 10 surrogate models range from 12.37\% to 16.93\% and the TPRs vary from 70.27\% to 78.35\%.
Last, among the 10 datasets training the top 10 surrogate ad targeting models, the percentage of positive data (indicating a high bid is triggered) is 11.26\% on average, ranging from 8.55\% to 15.38\%.

\begin{table}[!t]
\footnotesize
\centering
    \caption{Accuracy of surrogate user profiling and ad targeting models. FPR and TFP denote false positive and true positive rates.}
    \begin{tabular}{p{1.2in}<{\centering}p{1.in}<{\centering}p{1.in}<{\centering}}
    \hline
    Model type & User profiling &Ad targeting \\\hline
    Number of models & 20      & 10 \\
    Dataset size     & 10,000  & 10,000 \\
    Positive data    & 7.04\%  & 11.26\% \\
    Average FPR      & 3.92\%  & 18.28\%  \\
    Average TPR      & 96.57\% & 73.43\%  \\\hline
    \end{tabular}
    \label{tab:predictor}
\vspace{-.2in}
\end{table}

\vspace{-.05in}
\subsection{Baselines} 
\label{subsec: baselines}

We compare the performance of \ToolX against four other baseline approaches.
Two of these approaches (AdNauseam and TrackThis) have their own set of obfuscation URLs whereas the other two (Rand-intent and Bias-intent) use different selection techniques on the set of obfuscation URLs used by \ToolX, see \ref{subsec: personalurls} for more details on these sets of obfuscation URLs.
The four approaches are as follows:

\paratitle{AdNauseam.} 
Every time an obfuscation URL is needed, we uniformly randomly select one of the AdNauseam URLs. 


\paratitle{TrackThis.} 
Every time an obfuscation URL is needed, we uniformly randomly select one of the TrackThis URLs. 

\paratitle{Rand-intent.} 
Every time an obfuscation URL is needed, we uniformly randomly select one of the 193 intent URL subcategories and pick one URL from this subcategory at random.

\paratitle{Bias-intent.} 
Every time an obfuscation URL is needed, we randomly select one of the 193 intent URL subcategories with the probability proportional to the average reward triggered by URLs in this subcategory, and pick one URL from this subcategory uniformly randomly. 


It is noteworthy that the aforementioned AdNauseam and TrackThis baselines are not exactly the same as the original implementations. 
The original AdNauseam implementation clicks on ad URLs that are on the current page, making it hard to control the budget and diversity of obfuscation URLs. 
The original TrackThis implementation opens 100 preset URLs. 
We try to adapt these approaches to our experimental setup as best as possible. 
To this end, we first use the original implementations to generate the AdNauseam and TrackThis URL sets as described in the URL set subsection, and then randomly select obfuscation URLs from these sets with uniform probability, as already discussed. 

\begin{figure*}[!t]
    \centering
    \includegraphics[width=0.95\linewidth]{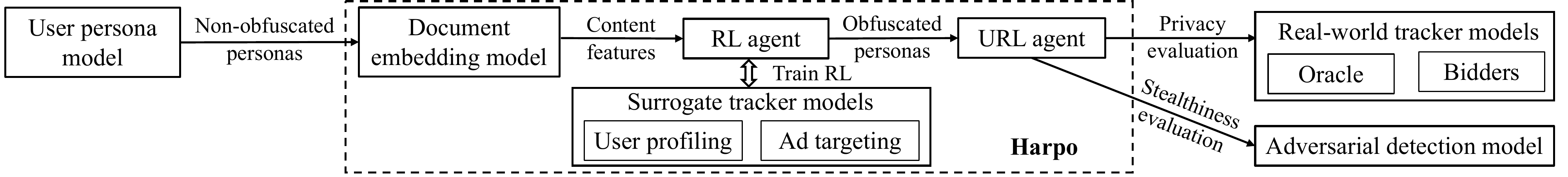}
    \caption{Overview of \ToolX's evaluation process.}
    \label{fig:models}
\vspace{-.2in}    
\end{figure*}

\section{Evaluation}
\label{sec: evaluation}
\F \ref{fig:models} summarizes \ToolX's evaluation process. 
We first use the user persona model of \S \ref{subsec: persona model} to generate a large number of diverse web browsing profiles. 
Next, we use the 4 \ToolX modules discussed in \S \ref{subsec:design} in the following order: (i) we use the doc2vec embedding model to extract  features for pages visited by each persona, (ii) we crawl data to train surrogate user profiling and ad targeting models and use them to train the RL agent, (iii) we use the RL agent to select obfuscation URLs, and (iv) we use the URL agent to create obfuscated personas. 
Then, we evaluate \ToolX's effectiveness in protecting user privacy as compared to the baselines against real-world user profiling and ad targeting models. 
Finally, we analyze \ToolX's performance from three key perspectives: overhead, stealthiness (using an adversarial detection model introduced later in \S \ref{subsec:stealth}), and adaptiveness.

\vspace{-.05in}
\subsection{Privacy}
\label{subsec:privacy}
\T \ref{tab:eval1} reports the effectiveness of \ToolX and baselines in protecting user privacy against surrogate user profiling ($L_1$ and $L_2$) and ad targeting ($L_3$ and $L_4$) models. 
Here, we only report the results for $\alpha=0.1$. 
Section \ref{subsec:budget} reports the results for varying values of $\alpha$.
Note that the Control represents a persona that does not deploy obfuscation.

\paratitle{User profiling.}
We note that \ToolX outperforms all four baselines with respect to both $L_1$ and $L_2$ metrics.
\ToolX triggers an average of 36.31\% ($L_1$) interest segments that were not present in the corresponding Control persona. The obfuscated persona has on average 4.40 ($L_2$) different interest segments from the corresponding Control persona, where on average 4.17 are new segments and 0.23 are removed segments from the Control persona.\footnote{While the vast majority of different segments between the Control and obfuscated persona are newly added segments here, when evaluating \ToolX against real-world user profiling and ad targeting models 25\% of different segments are due to removals, see \S \ref{subsection: transferability}.}
While Rand-intent and Bias-intent fare much better than AdNauseam and TrackThis, \ToolX outperforms all of the baselines by at least 1.41$\times$ and up to 2.92$\times$ in terms of $L_1$.
Similarly, \ToolX outperforms all baselines by at least 1.46$\times$ and up to 4.00$\times$ in terms of $L_2$.

\paratitle{Ad targeting.} 
We again note that \ToolX outperforms all four baseline approaches in terms of $L_3$ metric.\footnote{Note that we do not report results for $L_4$ because the surrogate model can only predict whether the bid is high or low, and not its actual value. We evaluate $L_4$ in Section \ref{subsection: transferability}.}
\ToolX increases high bids by 38.96\% as compared to the Control persona.
\ToolX again outperforms all of the baselines.
Bias-intent is the most competitive baseline triggering 24.72\% high bids on average. 
However, \ToolX is able to outperform it significantly by triggering 1.58$\times$ more high bids.

\begin{table}[!htbp]
\centering
\footnotesize
\caption{Evaluation results with surrogate models w.r.t.  
$L_1$ (percent of false segments in obfuscated persona),
$L_2$ (number of different segments between base and obfuscated persona),
$L_3$ (percentage increase of high bids in obfuscated persona),
$L_4$ (average ratio of obfuscated persona over base persona bid values).}
\vspace{-.1in}
\begin{tabular}{p{1in}<{\centering}p{0.55in}<{\centering}p{0.55in}<{\centering}p{0.55in}<{\centering}p{0.55in}<{\centering}}
\hline

\multirow{2}{*}{Approaches} & \multicolumn{4}{c}{Metrics}\\ \cline{2-5}
              & $L_1$            & $L_2$            & $L_3$          & $L_4$  \\ \hline
Control        & 0.00\%          & 0.00             & 0.00\%         &  --  \\
AdNauseam      & 12.42\%         & 1.10             & 2.78\%         &  --  \\
TrackThis      & 17.76\%         & 1.42             & 11.00\%        &  --  \\
Rand-intent    & 23.06\%         & 1.75             & 14.84\%        &  --  \\ 
Bias-intent    & 25.71\%         & 3.01             & 24.72\%        &  --  \\
\ToolX     & \textbf{36.31\%}   & \textbf{4.40} & \textbf{38.96\%} &  --  \\ \hline
\end{tabular}
\label{tab:eval1}
\vspace{-.15in}
\end{table}

\begin{table}[!t]
\label{tab:eval2}
\caption{Transferability results w.r.t.  
$L_1$ (percent of different segments between base and obfuscated profile),
$L_2$ (percent of false segments in obfuscated profile),
$L_3$ (percentage increase of high bids in obfuscated profile),
$L_4$ (average ratio of obfuscated persona over base persona bid values),
CPM (cost per thousand impressions in dollar, the unit of bid values).}
\vspace{-.1in}
\begin{subtable}{0.45\textwidth}
\centering
\footnotesize
\caption{Effectiveness against real-world tracker models used in training, with synthetic user personas as inputs}
\label{tab:eval2-1}
\begin{tabular}{p{1.2in}<{\centering}
p{0.5in}<{\centering}p{0.4in}<{\centering}p{0.5in}<{\centering}p{0.65in}<{\centering}}
\hline
\multirow{2}{*}{Approaches} & \multicolumn{4}{c}{Metrics}\\ \cline{2-5} 
               & $L_1$            & $L_2$          & $L_3$            & $L_4$ (CPM)  \\ \hline
Control        & 0.00\%           & 0.00           & 0.00\%           & 1.00 (\$0.29) \\
AdNauseam      & 12.85\%          & 1.53           & 2.70\%           & 1.21 (\$0.35) \\
TrackThis      & 32.67\%          & 2.81           & -1.50\%          & 0.89 (\$0.26) \\
Rand-intent    & 33.10\%          & 3.18           & 8.40\%           & 1.69 (\$0.49) \\ 
Bias-intent    & 31.27\%          & 3.19           & 10.30\%          & 2.07 (\$0.60) \\
\ToolX         & \textbf{43.24\%} & \textbf{5.22}  & \textbf{43.30\%} & \textbf{6.28 (\$1.82)}\\ \hline
\end{tabular}
\end{subtable}

\begin{subtable}{0.45\textwidth}
\centering
\footnotesize
\caption{Effectiveness against real-world tracker models not used in training, with synthetic user personas as inputs}
\label{tab:eval2-2}
\begin{tabular}{p{1.2in}<{\centering}
p{0.5in}<{\centering}p{0.4in}<{\centering}p{0.5in}<{\centering}p{0.65in}<{\centering}}
\hline
\multirow{2}{*}{Approaches} & \multicolumn{4}{c}{Metrics}\\ \cline{2-5} 
               & $L_1$            & $L_2$          & $L_3$            & $L_4$ (CPM)  \\ \hline
Bias-intent with $L1$        & --      & --  & 3.60\%  & 1.40 (\$0.40)\\
\ToolX with $L1$             & --      & --  & 10.20\% & 2.06 (\$0.59)\\
Bias-intent with $L2$        & -- & --    & 9.6\%   & 1.73 (\$0.50)\\
\ToolX with $L2$             & -- & --    & 10.10\% & 2.10 (\$0.61)\\
Bias-intent with $L3$        & 24.70\% & 2.55  & --      & --\\
\ToolX with $L3$             & 46.72\% & 2.97  & --      & --\\\hline
\end{tabular}
\end{subtable}
\begin{subtable}{0.45\textwidth}
\centering
\footnotesize
\caption{Effectiveness against real-world tracker models using real user personas from AOL dataset}
\label{tab:eval2-3}
\begin{tabular}{p{1.2in}<{\centering}
p{0.5in}<{\centering}p{0.4in}<{\centering}p{0.5in}<{\centering}p{0.7in}<{\centering}}
\hline
\multirow{2}{*}{Approaches} & \multicolumn{4}{c}{Metrics}\\ \cline{2-5} 
               & $L_1$            & $L_2$          & $L_3$            & $L_4$ (CPM)  \\ \hline
Control        & 0.00\%           & 0.00           & 0.00\%           & 1.00 (\$0.09) \\
AdNauseam      & 5.50\%          & 0.60           & 1.30\%           & 1.27 (\$0.15) \\
TrackThis      & 12.97\%          & 1.58           & 0.00\%           & 0.84 (\$0.11) \\
Rand-intent    & 24.24\%          & 2.24           & 0.90\%           & 2.22 (\$0.20) \\ 
Bias-intent    & 19.39\%          & 1.75           & 16.00\%          & 6.67 (\$0.60) \\
\ToolX         & \textbf{45.06\%} & \textbf{5.14}  & \textbf{49.10\%} & \textbf{18.96 (\$1.71)}\\ \hline
\end{tabular}
\end{subtable}

\vspace{-.2in}
\end{table}
\subsection{Transferability}
\label{subsection: transferability}
Next, we evaluate the effectiveness of \ToolX and baselines against real-world user profiling and ad targeting models. 
To this end, we replace surrogate models with the real-world user profiling model by Oracle Data Cloud Registry and ad targeting models of 10 different bidders. 
\T \ref{tab:eval2-1} reports the effectiveness of \ToolX and baselines against real-world user profiling ($L_1$ and $L_2$) and ad targeting ($L_3$ and $L_4$) models.

\paratitle{User profiling.} 
We again note that \ToolX outperforms all four baselines with respect to both $L_1$ and $L_2$ metrics, as shown in \T \ref{tab:eval2-1}. 
In fact, \ToolX's margin of improvement over baselines further increases against real-world models as compared to surrogate models. 
\ToolX now triggers an average of 43.24\% ($L_1$) interest segments that were not present in the corresponding Control persona. The obfuscated persona now has on average 5.22 ($L_2$) different interest segments from the corresponding Control persona, where 3.89 on average are new interest segments and 1.33 are removed segments from the Control persona.
\ToolX outperforms all baselines by at least 1.31$\times$ and up to 3.36$\times$ in terms of $L_1$.
Similarly, \ToolX outperforms all of the baselines by at least 1.64$\times$ and up to 3.41$\times$ in terms of $L_2$.

\paratitle{Ad targeting.} 
 As reported in \T \ref{tab:eval2-1}, \ToolX increases high bids by 43.30\% ($L_3$) and bid values by 6.28$\times$ ($L_4$) as compared to the Control persona.
We again note that \ToolX's margin of improvement over baselines further increases against real-world models as compared to surrogate models. 
\ToolX significantly outperforms all baselines by up to 16.04$\times$ in terms of $L_3$ and 7.06$\times$ in terms of $L_4$.
Bias-intent is again the most competitive baseline, but it increases high bids by only 10.30\% and bid values by only $2.07$. 
\ToolX is able to outperform it significantly by triggering 4.03$\times$ more high bids and 3.03$\times$ higher bid values. 

\vspace{-.05in}
\paratitle{Cross-validation against real-world tracker models.} 
Our transferability analysis so far has demonstrated that \ToolX's effectiveness against user profiling/ad targeting surrogate models can be transferred to the real-world user profiling/ad targeting models well. To further investigate \ToolX's transferability performance, we cross-validate \ToolX by testing it against different real-world tracker models than those used to train it.

\T \ref{tab:eval2-2} reports two type of results. In the first four rows, \ToolX is trained with user profiling models (w.r.t. $L_1$ or $L_2$) and tested against other models (e.g. against real-world ad targeting models, see $L_3$ and $L_4$ results). 
In the last two rows, \ToolX is trained with ad targeting models (w.r.t. $L_3$) and tested against real-world user profiling models (see $L_1$ and $L_2$ results).
As expected, its effectiveness is somewhat lower when it is tested against different models than the ones it was trained with, see \T \ref{tab:eval2-1} versus \ref{tab:eval2-2} results. 
That said, \ToolX performs well regardless. For example, it increases the average bid values by more than 2$\times$ when trained with user profiling models, and it creates obfuscated personas which have on average 2.97 different interest segments from the corresponding Control persona when trained with ad targeting models.
When comparing cross validation results for \ToolX and baselines (the table shows results only for Bias-intent as for the rest of the baselines the results do not change from those in \T \ref{tab:eval2-1}), when trained with user profiling models \ToolX outperforms baselines by up to 3.76$\times$ in terms of $L_3$ and 2.34$\times$ in terms of $L_4$ (i.e. against real-world ad targeting models). 
Similarly, when trained with ad targeting models, \ToolX outperforms all of the baselines against real-world user profiling models, by at least 1.17$\times$ and up to 2.79$\times$ in terms of $L_1$ and $L_2$ on average.

\paratitle{Evaluation using real user personas.} 
We have thus far evaluated \ToolX's effectiveness using synthetic user personas. 
Next, we evaluate \ToolX's effectiveness using real user personas. 
To this end, we randomly sample 100 real-world user personas from the AOL dataset and use them as non-obfuscated personas. 
Then, we use \ToolX and baselines approaches 
to generate 100 obfuscated personas and evaluate the effectiveness of  obfuscation against real-world tracker models.

\T \ref{tab:eval2-3} shows that \ToolX continues to significantly outperform all baselines.
Specifically, \ToolX outperforms all baselines against real-world user profiling models by up to 8.19$\times$ and 8.57$\times$ in terms of $L_1$ and $L_2$, respectively.
Also, \ToolX outperforms all baselines against real-world ad targeting models by up to 54$\times$ and 22.57$\times$ in terms of $L_3$ and $L_4$, respectively.
These results demonstrate that \ToolX under real personas achieves comparable results in terms of $L_1$ and $L_2$ and better results in terms of $L_3$ and $L_4$ as compared to under synthetic personas, which demonstrates \ToolX's transferability under real user personas.
Note that the CPM value for Control in Table \ref{tab:eval2-3} is lower than that for synthetic personas in Table \ref{tab:eval2-1} yielding a large gap between the value of $L_4$ under Table \ref{tab:eval2-1} and 
\ref{tab:eval2-3}, but the actual CPM value for \ToolX is comparable between the two tables.

In conclusion, our results demonstrate that \ToolX's performance transfers well to different real-world tracker models encountered in the wild as well as to real user personas.
The trends are largely consistent across surrogate and real-world models and across synthetic and real user personas.  
In fact, the performance gap between \ToolX and baselines widens in the real-world evaluation.
It is worth mentioning that real-world user profiling and ad targeting models may change over time.
While our results here demonstrate that \ToolX transfers well to real-world models, it might be prudent to update \ToolX from time to time to account for significant changes.  
We remark that \ToolX's RL agent is amenable to be updated in an online fashion and can also leverage transfer learning techniques to avoid training from scratch.

\begin{figure*}[!t]
\begin{subfigure}{.24\textwidth}
    \includegraphics[width=.95\linewidth]{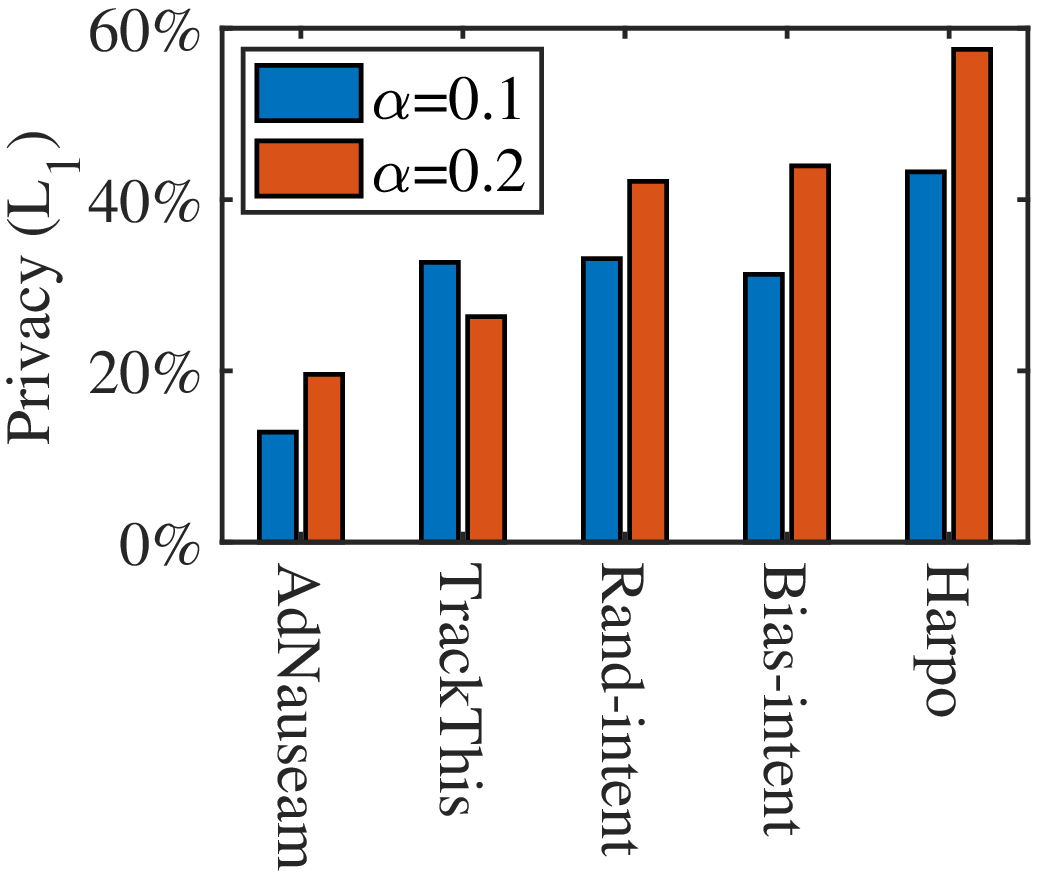}
    \caption{User profiling loss w.r.t. $L_1$}
    \label{fig:l1_budget}
\end{subfigure}
\begin{subfigure}{.24\textwidth}
    \includegraphics[width=.95\linewidth]{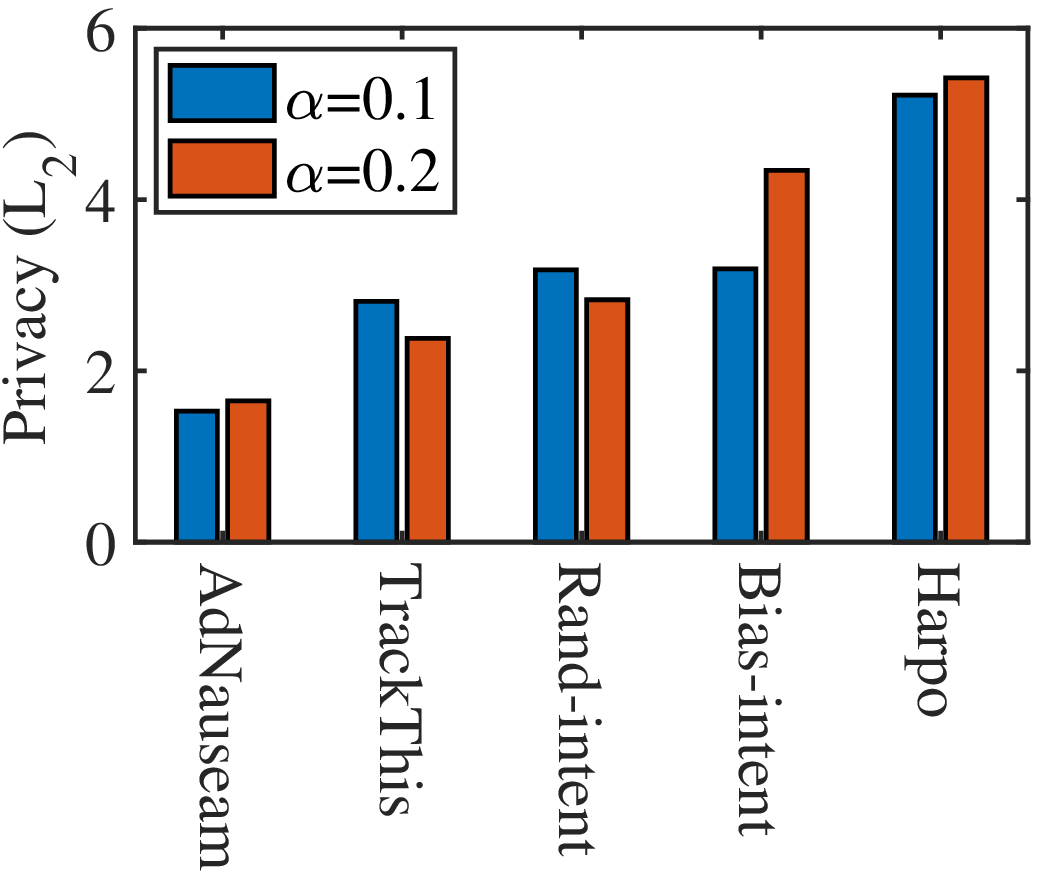}
    \caption{User profiling loss w.r.t. $L_2$}
    \label{fig:l2_budget}
\end{subfigure}
\begin{subfigure}{.24\textwidth}
    \includegraphics[width=.95\linewidth]{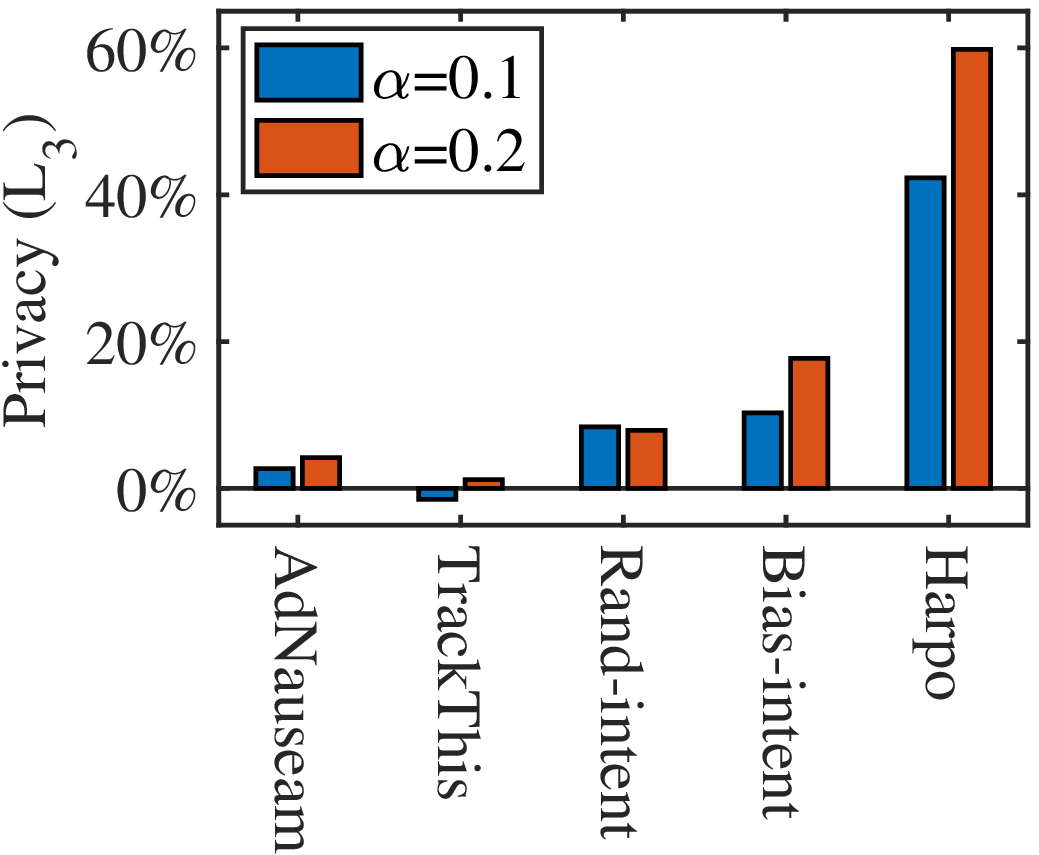}
    \caption{Ad targeting loss w.r.t. $L_3$}
    \label{fig:l3_budget}
\end{subfigure}
\begin{subfigure}{.24\textwidth}
    \includegraphics[width=.95\linewidth]{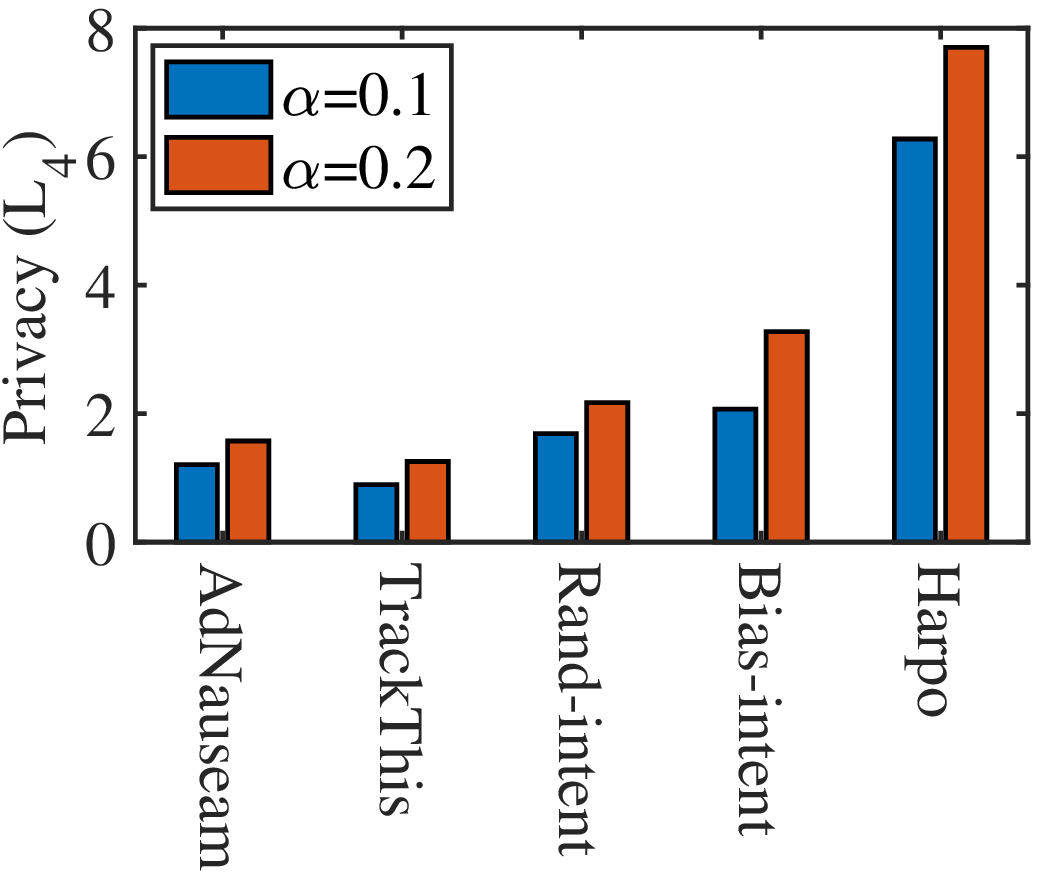}
    \caption{Ad targeting loss w.r.t. $L_4$}
    \label{fig:l4_budget}
\end{subfigure}
\vspace{-.1in}
\caption{Loss under different obfuscation budgets for the user profiling and ad targeting models. Note that the reported loss values ($L_1$, $L_2$, $L_3$) are all against real-world user profiling and ad targeting models. } 
\label{fig:budget}
\vspace{-.05in}
\end{figure*}

\subsection{Overhead}
\label{subsec:budget}
\paratitle{Obfuscation overhead.} Our evaluation thus far has used the obfuscation budget of $\alpha=0.1$.
Next, we investigate the impact of varying the obfuscation budget, controlled by the parameter $\alpha$, on the effectiveness of  \ToolX and baselines.
Figure \ref{fig:budget} plots the impact of varying $\alpha$ between $0.1$ and $0.2$ on real-world user profiling and ad targeting models. 
While there is a general increase in the effectiveness for a larger obfuscation budget, it is noteworthy that some baselines actually degrade when $\alpha$ is increased from  $0.1$ to $0.2$.
We note that \ToolX's effectiveness generally improves for the larger obfuscation budget and it continues to outperform the baselines. 
\ToolX's effectiveness improves by 1.33$\times$ for $L_1$,  1.03$\times$ for $L_2$, 1.41$\times$ for $L_3$, and 1.23$\times$ for $L_4$ when $\alpha$ is increased from $0.1$ to $0.2$.
In fact, \ToolX outperforms baselines even with a lower obfuscation budget.
Overall, \ToolX at $\alpha=0.1$ outperforms baselines at $\alpha=0.2$ by 
at least 1.47$\times$ in terms of $L_2$ on average and up to 13.27$\times$ in terms of $L_3$ on average.

\begin{figure*}[!t]
\centering
\begin{subfigure}{.32\textwidth}
    \includegraphics[width=.95\linewidth]{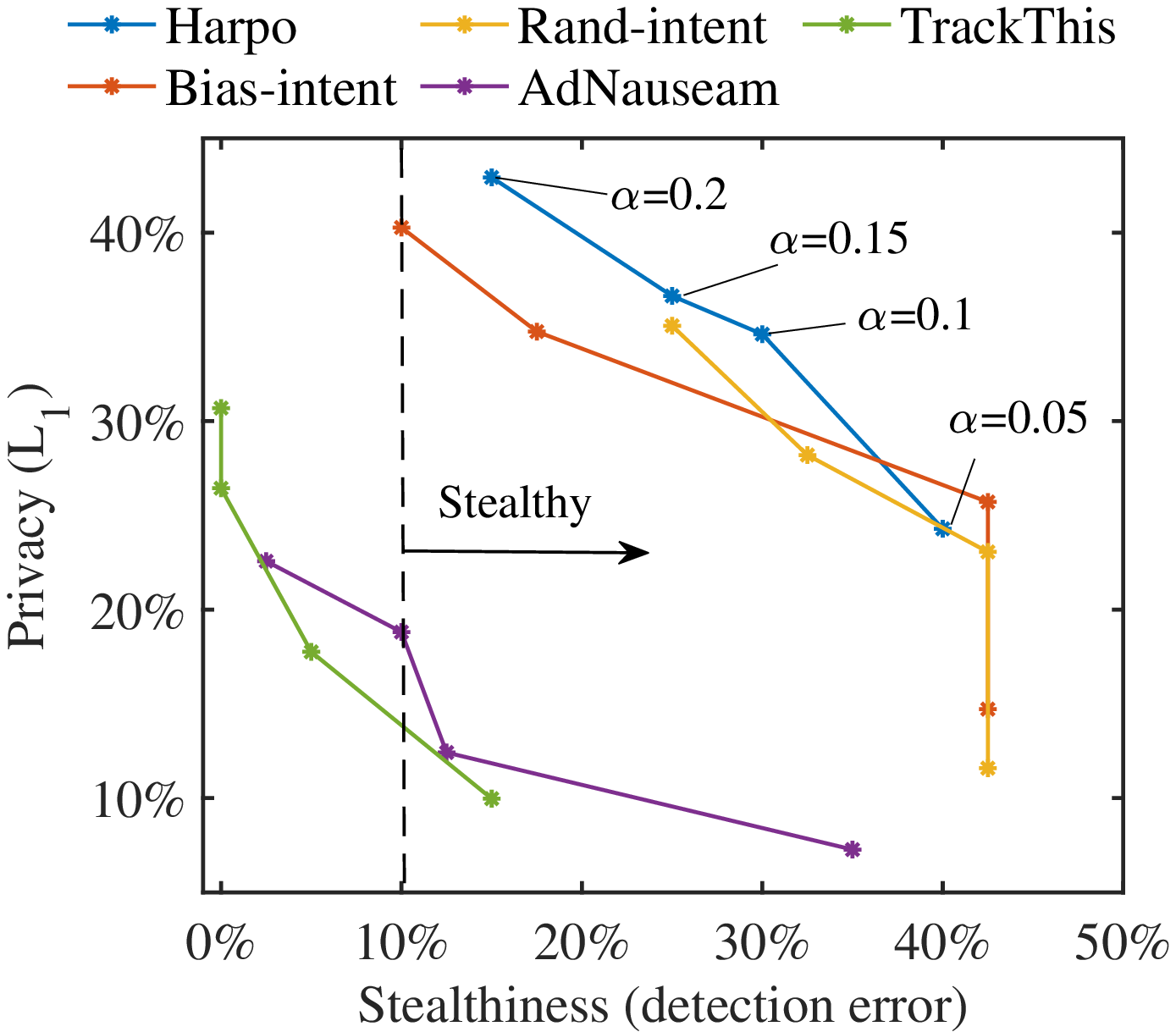}
    \caption{Privacy and stealthiness trade-off w.r.t. $L_1$}
    \label{fig:l1_stealth}
\end{subfigure}
\begin{subfigure}{.32\textwidth}
    \includegraphics[width=.95\linewidth]{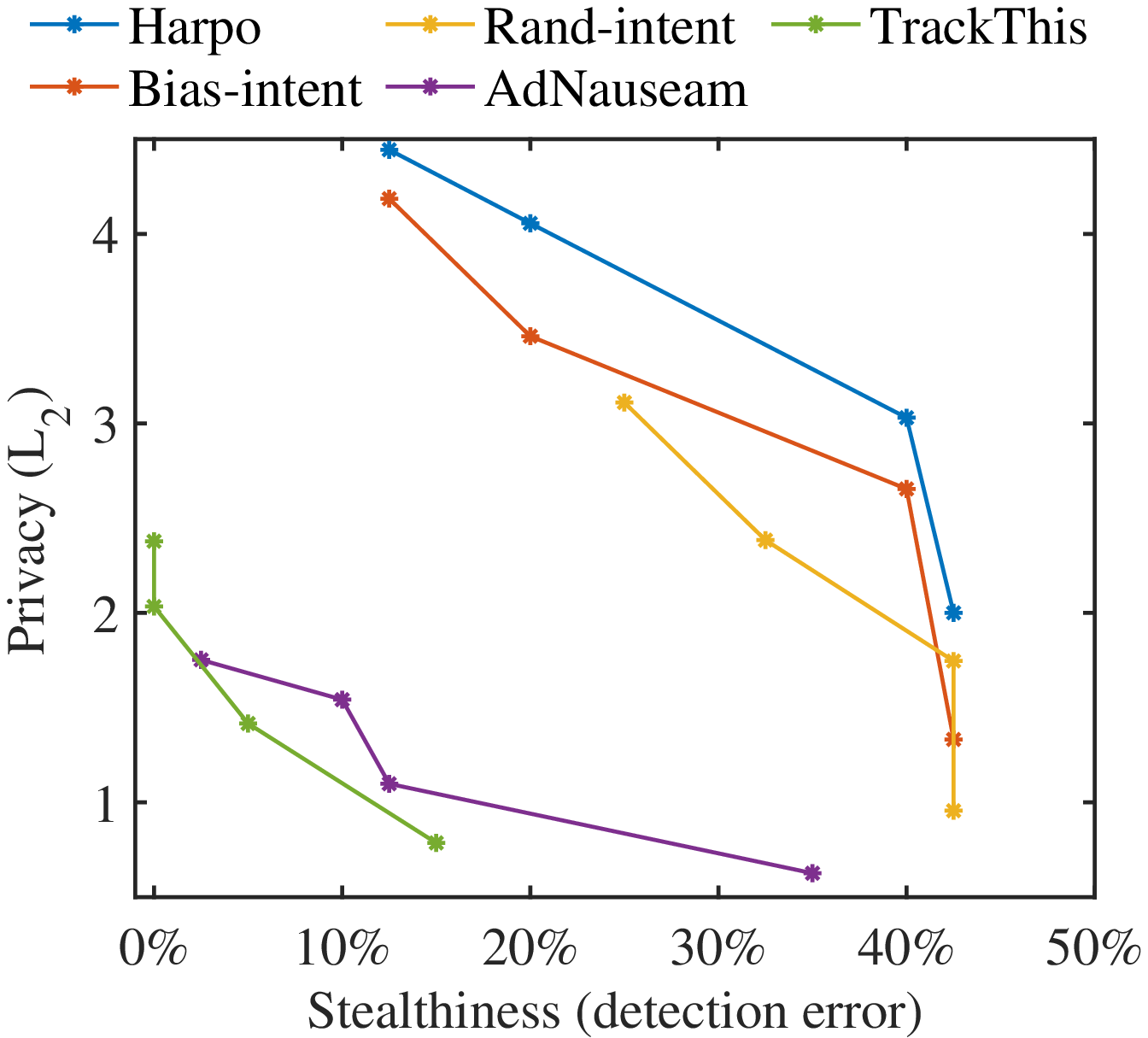}
    \caption{Privacy and stealthiness trade-off w.r.t. $L_2$}
    \label{fig:l2_stealth}
\end{subfigure}
\begin{subfigure}{.32\textwidth}
    \includegraphics[width=.95\linewidth]{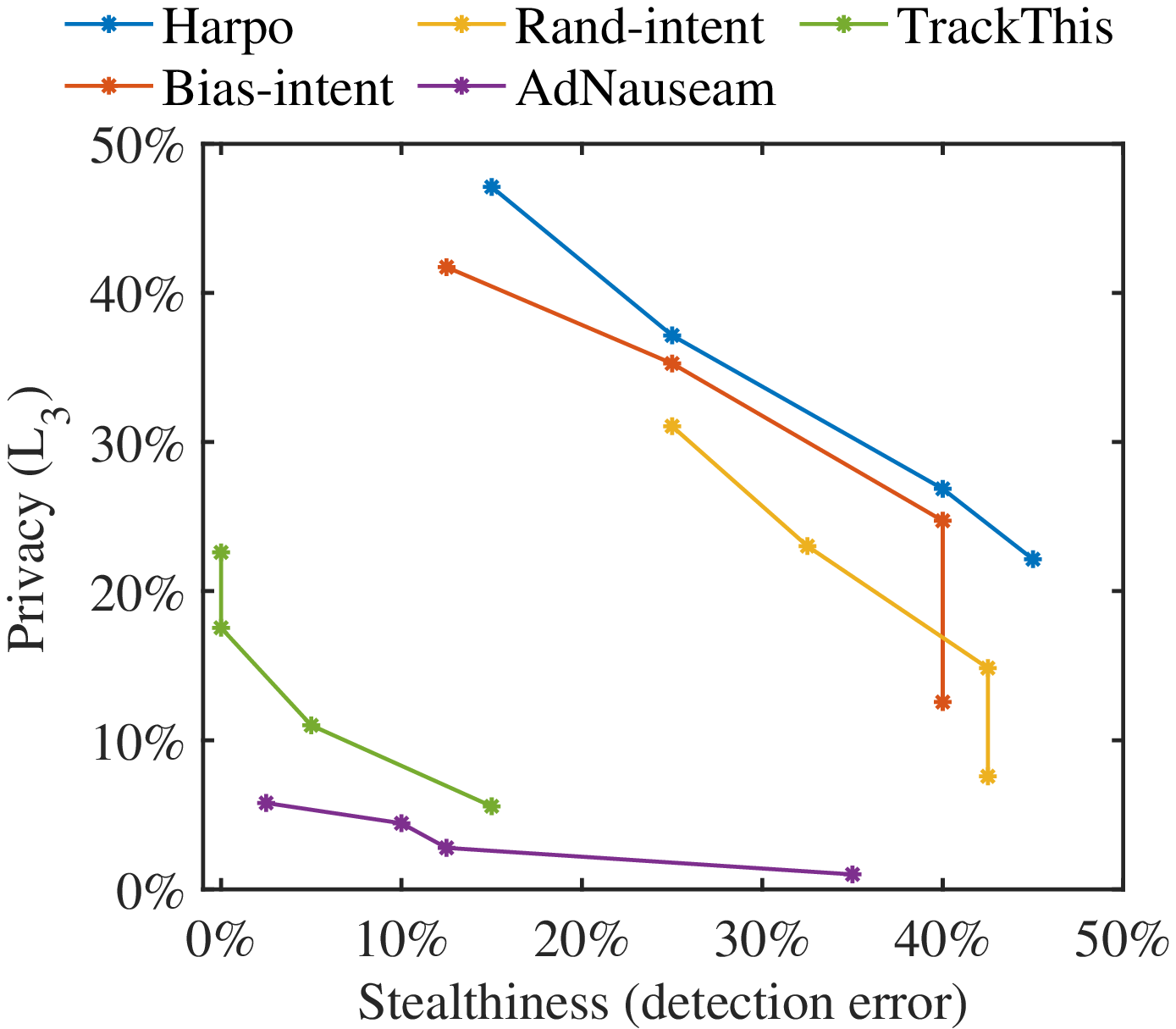}
    \caption{Privacy and stealthiness trade-off w.r.t. $L_3$}
    \label{fig:l3_stealth}
\end{subfigure}
\vspace{-.1in}
\caption{Stealthiness evaluation results. Note that the $\alpha$ values from left to right of each curve in each figure are $0.2$, $0.15$, $0.1$ and $0.05$ respectively. The reported privacy values ($L_1$, $L_2$, $L_3$) are against surrogate user profiling and ad targeting models.} 
\label{fig:stealth}
\vspace{-.2in}
\end{figure*}

\paratitle{System overhead.} 
We evaluate the system overhead of \ToolX{} to assess its potential adverse impact on user experience. 
We study \ToolX's system overhead in terms of resource consumption (CPU and memory usage) and overall user experience (page load time).
We launch a total of 300 browsing sessions on a commodity Intel Core i7 laptop with 8 GB memory on a residential WiFi network, without \ToolX{} as control and with \ToolX{} for $\alpha=$ 0.1 and 0.2.
Each browsing session involved visiting a series of 20 pages, with the next page being loaded as soon as the first page finished loading. 
For each page visit during the browsing session, we measure the CPU and memory consumption as well as the page load time.
The average percentage increase in CPU usage over control is 5.3\% and 8.8\% for $\alpha=$ 0.1 and 0.2, respectively.
The average percentage increase in memory usage over control is 3.9\% and 4.0\% for $\alpha=$ 0.1 and 0.2, respectively.
The average percentage increase in page load time over control is 0.20 and 0.26 seconds for $\alpha=$ 0.1 and 0.2, respectively.
We conclude that increasing values of $\alpha$ has a modest impact on the CPU and memory but a minimal impact on overall user experience. 
This is because \ToolX's browser extension implementation (model and fake page visits) uses a separate background thread that does not directly interrupt the browser's main thread. 
Overall, we expect \ToolX's implementation to have negligible system overheads on reasonably well-provisioned devices.


\subsection{Stealthiness}
\label{subsec:stealth}
Next, we introduce the notion of \textit{stealthiness} to reason about potential countermeasures by the tracker against \ToolX.
More specifically, we expect the tracker to try to detect the usage of \ToolX using purpose-built ML models.
We evaluate the stealthiness of \ToolX and baselines as well as study the trade-off between stealthiness and obfuscation budget ($\alpha$).

\paratitle{Adversarial detection.}
To build a supervised detection model, the tracker needs to gather training data comprising of both non-obfuscated and obfuscated browsing profiles. 
To this end, we assume a strong adversary that has access to sufficient non-obfuscated browsing profiles as well as black-box access to obfuscators (including \ToolX) that can be used to gather obfuscated browsing profiles.
To train the classification model, we assume that the tracker extracts embedding based content features for all the URLs in the available positive and negative labeled browsing profiles.
Thus, we assume that the tracker: (1) can track all the URLs in a user's browsing profile; (2) is able to extract content features for any URL that a user may visit; and (3) has sufficient resources to gather training data and train an ML-based supervised detection model.
Based on these assumptions, we design a binary ML classifier that uses the doc2vec embeddings as features of a user browsing profile and outputs a binary detection decision to indicate whether or not a given persona is obfuscated by \ToolX (or other obfuscators under consideration).
We gather a dataset of obfuscated and non-obfuscated personas containing a total of 20,000 URLs and use a similar 80-20 split to train and test this detector.
We then use the detection error as a metric to measure stealthiness--obfuscation is more/less stealthy if the detection error is higher/lower.


\paratitle{Privacy-stealthiness trade-off.} 
We evaluate privacy and stealthiness of \ToolX and baselines as we vary $\alpha \in \{0.05, 0.10, 0.15, 0.20\}$ in \F\ref{fig:stealth}.
We note that stealthiness generally degrades for larger values of $\alpha$.
As also shown in Section \ref{subsec:budget}, we again note that privacy generally improves for larger values of $\alpha$.
Thus, we get the privacy-stealthiness trade-off curve as $\alpha$ is varied. This trade-off is intuitive as the higher the obfuscation budget ($\alpha$), the higher the privacy. Additionally, it should be easier for the detector to identify the presence of obfuscation when $\alpha$ is higher, leading to lower stealthiness.
It is noteworthy that \ToolX achieves the best privacy-stealthiness trade-off (towards the top right of \F\ref{fig:stealth}) as compared to baselines.
More specifically, for the same level of stealthiness, \ToolX outperforms all baselines with respect to various privacy metrics. 
Similarly, for the same level of privacy, it achieves better stealthiness than baselines. 

\ToolX achieves both high privacy and stealthiness and is more stealthy than baselines because it ensures that obfuscation URLs are varied by disincentivizing the selection of same URLs and URL categories thanks to the way we have designed the reward of the RL agent and the corresponding MDP (see \S \ref{subsec:sysmodel}).
Note that by varying $\delta$, the adjustable parameter in the reward function which controls the diversity of URL selection, from 0.001 to 0.1, \ToolX may achieve a range of privacy and stealthiness results.
While Bias-intent and Rand-intent are in the same ballpark as \ToolX, we note that AdNauseam and TrackThis by far achieve the worst privacy-stealthiness trade-off (towards the bottom left of \F\ref{fig:stealth}).
AdNauseam is not stealthy because it always selects an ad URL, which perhaps stands out to the obfuscation detector. Note that this occurs in spite of the fact  that, to account for real world user behavior, we make sure non-obfuscated personas include 5\% of advertising URLs, thus there are ad URLs in both the original and obfuscated profiles.
Similarly, our TrackThis implementation randomly selects one of the obfuscation URLs from a curated set.\footnote{The original TrackThis implementation uses four fixed set of curated obfuscation URL sets, and selects one of those to injected all its $\approx 100$ URLs at the same time for obfuscation, which can be trivially detected.}

We conclude that \ToolX is able to achieve better privacy-stealthiness trade-off as compared to baselines.
This is in part because \ToolX is able to achieve better privacy for a given obfuscation budget due to its principled learning based approach.
To further provide insights into \ToolX, we next analyze obfuscation URLs selected by \ToolX and two of the most competitive baselines (Bias-intent and Rand-intent).

\begin{figure*}[t!]
    \centering
    \begin{subfigure}[t]{0.28\linewidth}
        \centering
        \includegraphics[width=0.9\linewidth]{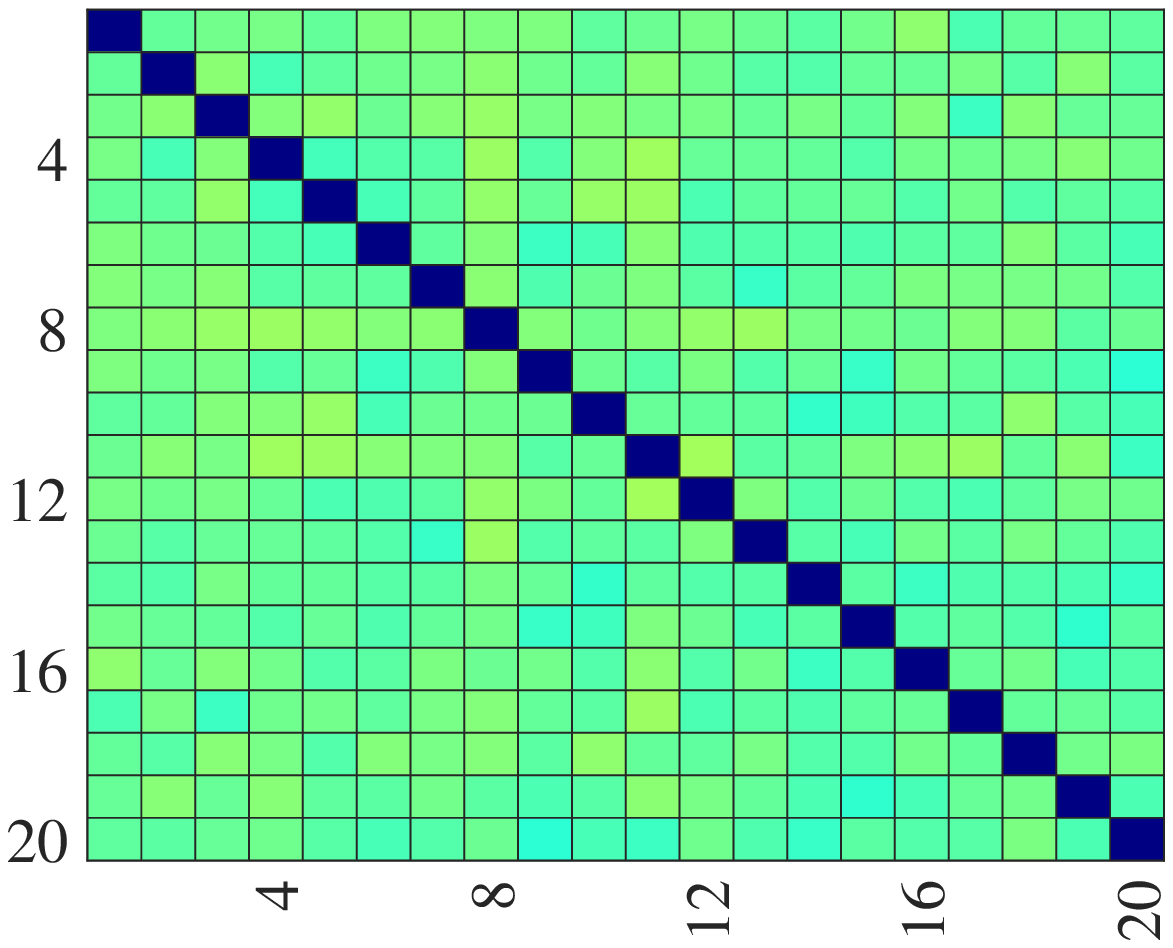}
         \caption{Adaptiveness of Rand-intent}
    \end{subfigure}%
    \begin{subfigure}[t]{0.28\linewidth}
        \centering
        \includegraphics[width=0.9\linewidth]{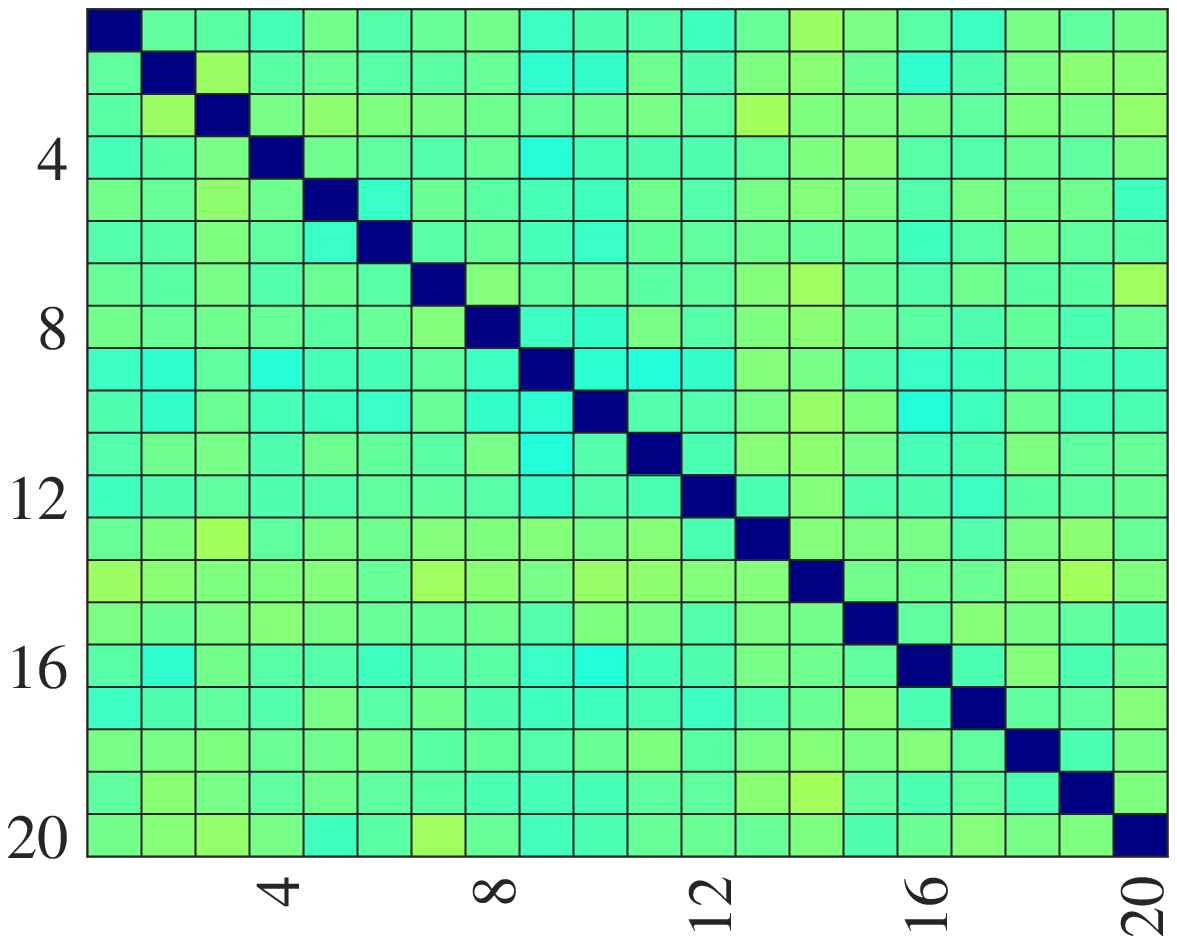}
         \caption{Adaptiveness of Bias-intent}
    \end{subfigure}
    \begin{subfigure}[t]{0.31\linewidth}
        \centering
        \includegraphics[width=0.9\linewidth]{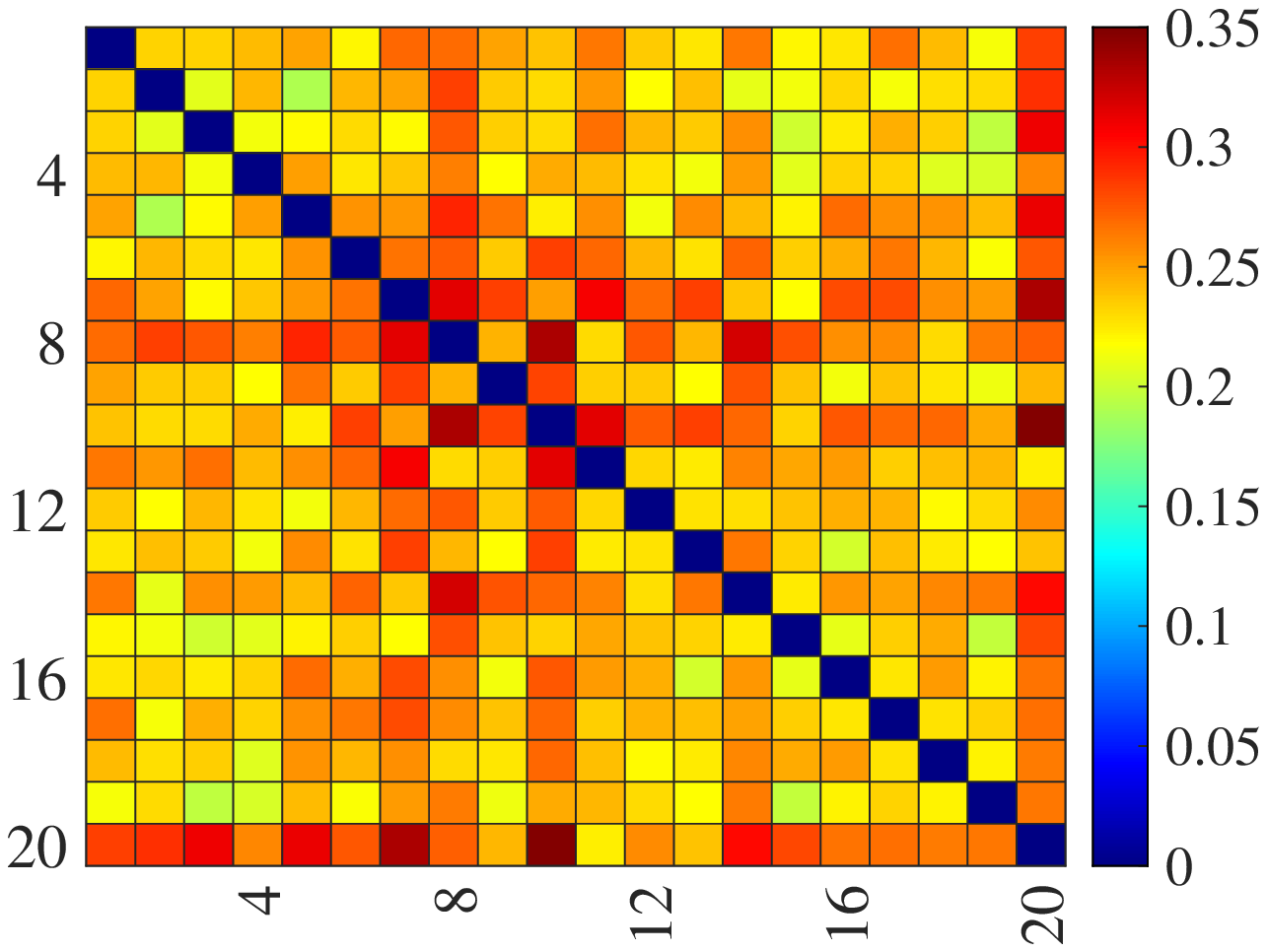}
         \caption{Adaptiveness of \ToolX}
    \end{subfigure}
        \vspace{-.1in}
    \caption{Adaptiveness of \ToolX and two of the most competitive baselines (Rand-intent and Bias-intent) against ad targeting models. The color of each cell represents the normalized Euclidean distance between a pair of obfuscation URL category distributions.
    Warmer colors (red; higher values) represent superior adaptiveness.    }
    \label{fig:adap}
    \vspace{-.2in}
\end{figure*}

\subsection{Adaptiveness}
\label{subsec:adap}
Our analysis thus far has not looked at whether and how the obfuscation URLs selected by \ToolX and baselines adapt to different user personas. 
To study \textit{adaptiveness} of obfuscation, we conduct a controlled experiment using a sample of 20 different personas (see Section \ref{subsec: persona model} for details).
To quantify the differences in selection of obfuscation URLs across each pair of personas, we visualize the distance between their distributions of obfuscation URL categories\footnote{Recall from Section \ref{subsec:trainingtesting} that \ToolX and baseline randomly select obfuscation URLs from a pool of 193 URL subcategories.} in \F \ref{fig:adap}.
Rows and columns here represent different persona types, and each cell in the matrix represents the normalized Euclidean distance between the corresponding pair of distributions of obfuscation URL categories.
If an obfuscation approach is not adaptive to different personas, we expect the values in the matrix to be closer to 0.
\F \ref{fig:adap} shows that \ToolX is clearly more adaptive across personas than our two most competitive baselines (Bias-intent and Rand-intent). 
The average adaptiveness of Bias-intent and Rand-intent is respectively $1.53\times$ and $1.51\times$ worse than \ToolX. 
Rand-intent and Bias-intent are less adaptive because they use a fixed distribution (uniform or weighted) to select obfuscation URL categories.
Overall, together with its principled nature, we believe that \ToolX's superior adaptiveness helps it achieve better privacy-stealthiness trade-off as compared to baselines.

\subsection{Personalization}
A user may
disallow \ToolX from distorting certain segments. 
This may be because the user wants to preserve a segment in his/her profile such that, for example, he/she may receive related ads \cite{chanchary2015user}. 
Or, it may be because the user does not want his/her profile to include a sensitive incorrect segment.
Motivated by this, we conducted an additional experiment where we trained \ToolX to distort \textit{allowed} segments while preserving \textit{disallowed} segments. 
Among the 20 considered interest segments, we select 15 as \textit{allowed} and 5 as \textit{disallowed}.

We denote the $L_2$ distortion on \textit{allowed} and \textit{disallowed} segments by $L_2^{allowed}$ and $L_2^{disallowed}$ respectively. 
Then, we train \ToolX to maximize $L_2^{allowed}-w_d L_2^{disallowed}$, i.e., to maximize the distortion in \textit{allowed} segments while minimizing the distortion in \textit{disallowed} segments. Note that $w_d$ is an adjustable parameter controlling how aggressive \ToolX is in distorting segments.\footnote{We set $w_d$ to 0.1 for this experiment.}
As shown in \T\ref{tab:eval5-1}, personalized \ToolX is able to trigger the same level of distortion for \textit{allowed} segments compared to non-personalized \ToolX. 
However, personalized \ToolX is able to preserve  \textit{disallowed} segments much better than non-personalized \ToolX. 
Such personalized obfuscation would provide \ToolX users more fine-grained control over their user profiles and subsequent ad targeting.

\begin{table}
\centering
\footnotesize
\caption{Personalization results. $L_2^{allowed}$ and $L_2^{disallowed}$ denote the distortion on \textit{allowed} segments and \textit{disallowed} segments respectively. Note that \ToolX is trained to maximize $L_2^{allowed} + L_2^{disallowed}$, while personalized \ToolX is trained to maximize $L_2^{allowed} - w_dL_2^{disallowed}$.}
\label{tab:eval5-1}
\begin{tabular}{p{1.3in}<{\centering}
p{1in}<{\centering}p{1in}<{\centering}}
\hline
                                 & $L_2^{allowed}$            & $L_2^{disallowed}$          \\ \hline
\ToolX                           & 3.74           & 0.71           \\
Personalized \ToolX              & 4.06           & 0.12         \\\hline
\end{tabular}
\vspace{-.1in}
\end{table}


\vspace{-.05in}
\section{Discussion}
\label{sec:ethics}

\subsection{Ethical Considerations}
We make a case that potential benefits of \ToolX to users outweigh potential harms to the online advertising ecosystem.

\noindent \textbf{Benefits to users.}
We argue that \ToolX meaningfully contributes to improving privacy for users who have no other recourse. 
The web's current business model at its core has been described as \textit{surveillance capitalism} \cite{zuboff2019age,cyphers2019behind,surveillance2019amnesty}.
The true extent of pervasive tracking and surveillance for the sake of online targeted advertising is unknown to lay users, who cannot be reasonably expected to understand the details buried in incomprehensible privacy policies \cite{amos2020privacy,vail2008empirical} or make the informed choices due to deceptive practices \cite{mathur2019dark}.
A vast majority of online advertising platforms do not support privacy-by-design features or attempt to circumvent privacy-enhancing blocking tools.
Thus, the practice of falsification in a privacy-enhancing obfuscation tool, such as \ToolX, is ethically justified from the user's perspective  \cite{brunton2013political}.


\noindent \textbf{Harms to online advertising ecosystem.}
We argue that potential harms of \ToolX to online advertising ecosystem are lower than existing obfuscation approaches or other alternatives. 
\ToolX does introduce additional costs/overheads for publishers and advertisers.
Since \ToolX reduces the effectiveness of user profiling and ad targeting, advertisers may have to spend more on advertising to achieve the same level of conversions. 
Publishers, in turn, may also notice a reduction in their advertising revenues. 
In the worst case where behavioral targeting is completely ineffective, advertisers may have to resort to contextual advertising that is reportedly about 52\% less valuable \cite{Johnson2019ConsumerPrivacyOpt,deepak2019effect}.  
However, we note that obfuscation is more viable as compared to other alternatives such as ad/tracker blocking, where advertisers and publishers essentially lose all advertising revenue.
Moreover, unlike AdNauseam, \ToolX is designed to not explicitly click on ads, thereby not engaging in overt ad fraud.

Thus, we argue that \ToolX provides an ecologically viable (though less profitable) path for the online advertising ecosystem while providing clear privacy benefits to users. 


\vspace{-.05in}
\subsection{Limitations}
\label{subsec:limitation}
We discuss some of the limitations of \ToolX's design, implementation, and evaluation.

\noindent \textbf{Side-channels.} 
There are side-channels that can be used to undermine \ToolX{}'s stealthiness.
Specifically, since \ToolX{} uses background tabs to load obfuscation URLs, an adversary can use the Page Visibility API \cite{URL_PAGEVISIBILITYAPI} or timing information via  Performance API \cite{URL_CHROMIUMTHROTTLE} to determine whether the tab is open in the background and detect the use of \ToolX{}.
More generally, \ToolX{}'s browser extension implementation is susceptible to extension fingerprinting attacks \cite{laperdrix2021fingerprinting,karami2020carnus,starov2017xhound,sjosten2017discovering}.
\ToolX{}'s implementation can be hardened against these attacks by patching the APIs that leak such information.

\noindent \textbf{Tracking modalities.}
Trackers use multiple modalities (browsing profile, page interactions, location, sensors, etc.) to profile user interests and subsequently target ads. 
While \ToolX{} is currently designed to only obfuscate users' browsing profiles, it can be extended to obfuscate other modalities in the future.



\noindent \textbf{User traces.}
%
We evaluated \ToolX using both real user traces from the 15 year old AOL dataset and synthetic traces based on our user persona model.
We acknowledge that the characteristics of the AOL and synthetic traces might be different than those of current real users. 
A future line of research would be to evaluate \ToolX{} by recruiting real users.

\section{Related Work}
\label{sec: related}

In this section, we contextualize our work with respect to prior literature on enhancing user privacy in online behavioral advertising.
Online advertising platforms typically do not allow users to meaningfully opt in/out of tracking. 
The notable exception is Apple's newly introduced App Tracking Transparency feature that requires apps to get permission from users to track them for targeted advertising \cite{APPLE_ATT}.
Unfortunately, a vast majority of data brokers do not give users any meaningful choice about tracking.
Thus, as we discuss next, the privacy community has developed a number of privacy-enhancing blocking and obfuscation tools geared towards online behavioral advertising.

\subsection{Privacy-enhancing blocking tools}
The privacy community has a long history of developing privacy-enhancing tools to counter online advertising and tracking through blocking.
These blocking tools have seen widespread adoption, with hundreds of millions of users across uBlock Origin \cite{URL_UBLOCKORIGIN}, AdBlock Plus \cite{URL_ABP}, and Ghostery \cite{URL_GHOSTERY}.
In fact, several privacy-focused browsers such as Firefox \cite{URL_FIREFOX} and Brave \cite{URL_BRAVE} now provide built-in blocking features.
An established line of research aims to improve the effectiveness of these blocking tools \cite{storey2017future,iqbal2018adgraph,gugelmann2015automated,bau2013promising,ikram2017towards,shuba2018nomoads,wu2016machine,yu2016tracking}.
In addition to blanket blocking of advertising and/or tracking, selective blocking tools aim to give users control over the trade-off between privacy and utility.
Tools such as MyTrackingChoices \cite{achara2016mytrackingchoices} and TrackMeOrNot \cite{meng2016trackmeornot} enable users to block tracking of private interests while allowing tracking of non-private interests. 
%
Thus, these selective blocking tools can help users still receive personalized ads for non-private interests while protecting their private interests.

Unsurprisingly, advertisers and trackers consider blocking a threat to their business model.
The ensuing arms race over the last few years has seen advertisers and trackers leveraging a myriad of ways to circumvent blocking \cite{iqbal2018adgraph,wang2016webranz,alrizah2019errors,chendetecting}.
First, blocking tools that rely on signatures (\eg EasyList) can be trivially evaded by simply modifying the signature (\eg randomizing domain names or URL paths) \cite{alrizah2019errors}.
Second, new circumvention techniques to bypass blocking techniques are often devised before being eventually patched \cite{daocharacterizing,bashir2018tracking,subramani2020measuring}.
Finally, prior work has demonstrated the non-stealthy nature of most forms of ad blocking as they can be reliably detected by publishers, allowing them to retaliate using anti-adblocking paywalls \cite{mughees2017detecting, nithyanand2016adblocking}.
Thus, blocking is not the silver bullet against online advertising and tracking.

\subsection{Privacy-enhancing obfuscation tools}
Closer to our focus in this paper, the privacy community has also developed privacy-enhancing obfuscation tools to counter online advertising and tracking \cite{finnn2015obfuscationbook}. 
We discuss prior obfuscation approaches in terms of whether the obfuscation approach is: (1) adaptive to the user's browsing profile, (2) principled in attacking the tracker's profiling/targeting model, (3) stealthy against detection and potential countermeasures by the tracker, and (4) cognizant of obfuscation overheads.

In a seminal work, Howe and Nissenbaum \cite{howe2017engineering} presented AdNauseam that combined blocking with obfuscation to ``protest'' against online targeted advertising.
The main aim is to protect user privacy by confusing user profiling and ad targeting systems used in online targeted advertising.
To this end, AdNauseam obfuscates a user's browsing behavior by deliberately clicking on a controllable fraction of encountered ads.
While AdNauseam's obfuscation approach is adaptive to the user's browsing and allows control of overheads, it is not principled and stealthy---it injects a random subset of ad URLs in a user's browsing profile without any awareness of the user profiling or ad targeting model.
In the same vein, Mozilla recently launched TrackThis \cite{URL_TRACKTHIS} to ``throw off'' advertisers and trackers by injecting a curated list of obfuscation URLs.
TrackThis is more primitive than AdNauseam---it is further not adaptive or stealthy because it injects a fixed set of curated obfuscation URLs that do not change across different user browsing profiles.

In an early work that does not specifically focus on online targeted advertising, Xing et al. \cite{xing2013takethis} proposed an attack to ``pollute'' a user's browsing profile and impact first-party personalization on YouTube, Google, and Amazon. 
Building on this work, Meng et al. \cite{ming2014pwned} implemented and deployed this polluting attack against online targeted advertising. 
Their obfuscation approach randomly injects curated URLs that are likely to trigger re-targeting.
In another attack on online targeted advertising, Kim et al. \cite{kim2018adbudgetkiller} proposed to create fake browsing profiles to waste an advertiser's budget on fake ad slots.
While similar to Meng et al. \cite{ming2014pwned} in that they aim to trigger more expensive re-targeted ads, their attack does not seek to enhance user privacy and is squarely focused on wasting the budget of advertisers.
While these obfuscating approaches were shown to impact ad targeting, they share the same limitations as TrackThis.

Degeling and Nierhoff \cite{Degeling18bluekaiwpes} designed and evaluated an obfuscation approach to ``trick'' a real-world user profiling system. 
While their obfuscation approach injects a curated set of obfuscation URLs, it is principled because it relies on feedback from the advertiser's user profiling model to select obfuscation URLs. 
Their obfuscation approach was shown to induce incorrect interest segments in BlueKai's user profiling model.
While their obfuscation approach is principled and somewhat adaptive to a user's browsing profile, it is neither stealthy nor cognizant of obfuscation overheads.



In a related obfuscation-through-aggregation approach, Biega et al. \cite{biega2017privacy} proposed to use a proxy to interleave browsing profiles of multiple users to protect their privacy through ``solidarity.''
Their approach mixes browsing profiles of different users  based on the similarity between their browsing profiles.
Their approach is adaptive and stealthy because it tries to mix browsing profiles of similar users. 
However, it is neither principled nor it is cognizant of obfuscation overheads.


Beigi et al. \cite{beigi2018protectinguserprivacy} proposed to use greedy search to suitably obfuscate a user's browsing profile. 
Their approach is adaptive and principled since it uses a greedy search approach that is essentially equivalent to our Bias-intent baseline.
However, it does not consider sequential dependencies \cite{chierichetti2012arewebusersmarkovian} in a user's browsing profile or allow control over obfuscation overheads.

%

%
%


\section{Conclusion}
\label{sec:conclusion}


In this paper we presented \ToolX, a principled reinforcement learning-based obfuscation approach to subvert online targeted advertising.
\ToolX significantly outperforms existing obfuscation tools by as much as 16$\times$ for the same overhead.
Additionally, for the same level of privacy, \ToolX provides better stealthiness against potential countermeasures.
Thus, the privacy protections offered by \ToolX are better suited for the arms race than existing obfuscation tools.
%
We hope that \ToolX and follow-up research will lead to a new class of obfuscation-driven effective, practical, and long lasting privacy protections against online behavioral advertising. 
To facilitate follow-up research, \ToolX's source code is available at \url{https://github.com/bitzj2015/Harpo-NDSS22}.


\section*{Acknowledgment}

The authors would like to thank John Cook for his help with initial data collection. This work is supported in part by the Robert N. Noyce Trust and the National Science Foundation under grant numbers 1956435, 1901488, 2051592, and 2103439.

\balance
\bibliographystyle{unsrt}
\bibliography{refs}

\begin{thebibliography}{100}

\bibitem{acar2014web}
Gunes Acar, Christian Eubank, Steven Englehardt, Marc Juarez, Arvind Narayanan,
  and Claudia Diaz.
\newblock {The Web Never Forgets: Persistent Tracking Mechanisms in the Wild}.
\newblock In {\em ACM Conference on Computer and Communications Security
  (CCS)}, 2014.

\bibitem{iqbal2020fingerprinting}
Umar Iqbal, Steven Englehardt, and Zubair Shafiq.
\newblock Fingerprinting the fingerprinters: Learning to detect browser
  fingerprinting behaviors.
\newblock {\em IEEE Symposium on Security \& Privacy (S\&P)}, 2021.

\bibitem{papadopoulos2019cookie}
Panagiotis Papadopoulos, Nicolas Kourtellis, and Evangelos Markatos.
\newblock Cookie synchronization: Everything you always wanted to know but were
  afraid to ask.
\newblock In {\em The World Wide Web Conference (WWW)}, 2019.

\bibitem{Englehardt161MillionSite}
Steven Englehardt and Arvind Narayanan.
\newblock {Online Tracking: A 1-million-site Measurement and Analysis}.
\newblock In {\em ACM Conference on Computer and Communications Security
  (CCS)}, 2016.

\bibitem{englehardt2015cookies}
Steven Englehardt, Dillon Reisman, Christian Eubank, Peter Zimmerman, Jonathan
  Mayer, Arvind Narayanan, and Edward~W Felten.
\newblock Cookies that give you away: The surveillance implications of web
  tracking.
\newblock In {\em International Conference on World Wide Web (WWW)}, 2015.

\bibitem{Olejnik2014SellingOff}
Lukasz Olejnik, Minh-Dung Tran, and Claude Castelluccia.
\newblock Selling off privacy at auction.
\newblock In {\em Network and Distributed Systems Security (NDSS) Symposium},
  2014.

\bibitem{federal2014databroker}
Federal Trade~Commission.
\newblock {Data brokers: A call for transparency and accountability}.
\newblock 2014.

\bibitem{kim2018jcr}
Tami Kim, Kate Barasz, and Leslie~K John.
\newblock {Why Am I Seeing This Ad? The Effect of Ad Transparency on Ad
  Effectiveness}.
\newblock {\em Journal of Consumer Research}, 45(5):906--932, 05 2018.

\bibitem{dehling2019consumer}
Tobias Dehling, Yuchen Zhang, and Ali Sunyaev.
\newblock Consumer perceptions of online behavioral advertising.
\newblock In {\em 2019 IEEE 21st Conference on Business Informatics (CBI)}.
  IEEE, 2019.

\bibitem{ur2012smart}
Blase Ur, Pedro~Giovanni Leon, Lorrie~Faith Cranor, Richard Shay, and Yang
  Wang.
\newblock Smart, useful, scary, creepy: perceptions of online behavioral
  advertising.
\newblock In {\em Symposium on Usable Privacy and Security (SOUPS}, 2012.

\bibitem{zuboff2019age}
Shoshana Zuboff.
\newblock {\em The Age of Surveillance Capitalism: The Fight for a Human Future
  at the New Frontier of Power}.
\newblock 2019.

\bibitem{URL_GOOGLEPRECOOKIES}
{Types of cookies used by Google}.
\newblock {\url{https://policies.google.com/technologies/types}}.

\bibitem{WAPO2013MobileInterceptTraffic}
Ashkan Soltani and Barton Gellman.
\newblock {New documents show how the NSA infers relationships based on mobile
  location data}.
\newblock {The Washington Post}, 2013.

\bibitem{URL_RFC7258}
S.~Farrell and H.~Tschofenig.
\newblock {Pervasive Monitoring Is an Attack}.
\newblock IETF RFC 7258, 2014.

\bibitem{NYTimes2018BiasWomen}
Noam Scheiber.
\newblock Facebook accused of allowing bias against women in job ads.
\newblock {The New York Times}, 2018.

\bibitem{NewScience2019FBOutsGayMen}
Chris Stokel-Walker.
\newblock Facebook's ad data may put millions of gay people at risk.
\newblock New Scientist, 2019.

\bibitem{Propublica2016FBRace}
Julia Angwin and Terry~Parris Jr.
\newblock Facebook lets advertisers exclude users by race.
\newblock ProPublica, 2016.

\bibitem{Matz2019Pnas}
S.~C. Matz, M.~Kosinski, G.~Nave, and D.~J. Stillwell.
\newblock Psychological targeting as an effective approach to digital mass
  persuasion.
\newblock {\em Proceedings of the National Academy of Sciences},
  114(48):12714--12719, 2017.

\bibitem{medium2018Ethics}
Dan Gizzi.
\newblock The ethics of political micro-targeting, 2018.

\bibitem{CNN2019FBIAds}
Donie O'Sullivan and David Shortell.
\newblock Exclusive: The fbi is running facebook ads targeting russians in
  washington.
\newblock 2019.

\bibitem{Intercept2016DeRadicalize}
Naomi LaChance.
\newblock Program to deradicalize jihadis will be used on right-wingers.
\newblock The Intercept, 2018.

\bibitem{APPLE_ATT}
{App Tracking Transparency}.
\newblock
  \url{https://developer.apple.com/documentation/apptrackingtransparency}.

\bibitem{URL_UBLOCKORIGIN}
Raymond Hill.
\newblock {An efficient blocker for Chromium and Firefox. Fast and lean, uBlock
  Origin}.
\newblock {\url{https://github.com/gorhill/uBlock\#ublock-origin}}, 2019.

\bibitem{URL_GHOSTERY}
{Ghostery.}
\newblock \url{https://www.ghostery.com/}.

\bibitem{iqbal2018adgraph}
Umar Iqbal, Peter Snyder, Shitong Zhu, Benjamin Livshits, Zhiyun Qian, and
  Zubair Shafiq.
\newblock Adgraph: A graph-based approach to ad and tracker blocking.
\newblock {\em IEEE Symposium on Security \& Privacy (S\&P)}, 2020.

\bibitem{wang2016webranz}
Weihang Wang, Yunhui Zheng, Xinyu Xing, Yonghwi Kwon, Xiangyu Zhang, and
  Patrick Eugster.
\newblock Webranz: Web page randomization for better advertisement delivery and
  web-bot prevention.
\newblock In {\em ACM International Symposium on Foundations of Software
  Engineering (FSE)}, 2016.

\bibitem{merzdovnik2017block}
Georg Merzdovnik, Markus Huber, Damjan Buhov, Nick Nikiforakis, Sebastian
  Neuner, Martin Schmiedecker, and Edgar Weippl.
\newblock Block me if you can: A large-scale study of tracker-blocking tools.
\newblock In {\em 2017 IEEE European Symposium on Security \& Privacy (Euro
  S\&P)}, 2017.

\bibitem{alrizah2019errors}
Mshabab Alrizah, Sencun Zhu, Xinyu Xing, and Gang Wang.
\newblock Errors, misunderstandings, and attacks: Analyzing the crowdsourcing
  process of ad-blocking systems.
\newblock In {\em ACM Internet Measurement Conference (IMC)}, 2019.

\bibitem{chendetecting}
Quan Chen, Peter Snyder, Ben Livshits, and Alexandros Kapravelos.
\newblock Detecting filter list evasion with event-loop-turn granularity
  javascript signatures.
\newblock In {\em IEEE Symposium on Security \& Privacy (S\&P)}, 2021.

\bibitem{vastel2018filters}
Peter Snyder, Antoine Vastel, and Ben Livshits.
\newblock Who filters the filters: Understanding the growth, usefulness and
  efficiency of crowdsourced ad blocking.
\newblock {\em Proceedings of the ACM on Measurement and Analysis of Computing
  Systems}, 4(2):1--24, 2020.

\bibitem{le21cvinspector}
Hieu Le, Athina Markopoulou, and Zubair Shafiq.
\newblock {CV-Inspector: Towards Automating Detection of Adblock
  Circumvention}.
\newblock In {\em Proceedings of the Symposium on Network and Distributed
  System Security (NDSS)}, 2021.

\bibitem{bashir2018tracking}
Muhammad~Ahmad Bashir, Sajjad Arshad, Engin Kirda, William Robertson, and
  Christo Wilson.
\newblock How tracking companies circumvented ad blockers using websockets.
\newblock In {\em ACM Internet Measurement Conference (IMC)}, 2018.

\bibitem{daocharacterizing}
Ha~Dao, Johan Mazel, and Kensuke Fukuda.
\newblock {Characterizing CNAME Cloaking-Based Tracking on the Web}.
\newblock In {\em Traffic Measurement and Analysis Conference (TMA)}, 2020.

\bibitem{subramani2020measuring}
Karthika Subramani, Xingzi Yuan, Omid Setayeshfar, Phani Vadrevu, Kyu~Hyung
  Lee, and Roberto Perdisci.
\newblock Measuring abuse in web push advertising.
\newblock {\em arXiv:2002.06448}, 2020.

\bibitem{howe2017engineering}
Daniel~C Howe and Helen Nissenbaum.
\newblock {Engineering Privacy and Protest: A Case Study of AdNauseam}.
\newblock In {\em International Workshop on Privacy Engineering (IWPE)}, 2017.

\bibitem{URL_TRACKTHIS}
{Hey advertisers, track THIS}.
\newblock \url{https://blog.mozilla.org/firefox/hey-advertisers-track-this},
  2019.

\bibitem{URL_ADNAUSEAM}
{AdNauseam - Clicking Ads So You Don't Have To}.
\newblock \url{https://adnauseam.io/}.

\bibitem{Degeling18bluekaiwpes}
Martin Degeling and Jan Nierhoff.
\newblock Tracking and tricking a profiler: Automated measuring and influencing
  of bluekai’s interest profiling.
\newblock In {\em Workshop on Privacy in the Electronic Society (WPES)}, 2018.

\bibitem{biega2017privacy}
Asia~J. Biega, Rishiraj Saha~Roy, and Gerhard Weikum.
\newblock Privacy through solidarity: A user-utility-preserving framework to
  counter profiling.
\newblock SIGIR '17, page 675–684, New York, NY, USA, 2017. Association for
  Computing Machinery.

\bibitem{beigi2018protectinguserprivacy}
Ghazaleh Beigi, Ruocheng Guo, Alexander Nou, Yanchao Zhang, and Huan Liu.
\newblock Protecting user privacy: An approach for untraceable web browsing
  history and unambiguous user profiles.
\newblock {\em CoRR}, abs/1811.09340, 2018.

\bibitem{URL_ORACLE}
{Oracle Data Cloud Registry Information}.
\newblock \url{https://datacloudoptout.oracle.com}.

\bibitem{Pachilakis2019headerbiddingimc}
Michalis Pachilakis, Panagiotis Papadopoulos, Evangelos~P. Markatos, and
  Nicolas Kourtellis.
\newblock {No More Chasing Waterfalls: A Measurement Study of the Header
  Bidding Ad-Ecosystem}.
\newblock In {\em ACM Internet Measurement Conference (IMC)}, 2019.

\bibitem{nikiforakis2013cookieless}
Nick Nikiforakis, Alexandros Kapravelos, Wouter Joosen, Christopher Kruegel,
  Frank Piessens, and Giovanni Vigna.
\newblock Cookieless monster: Exploring the ecosystem of web-based device
  fingerprinting.
\newblock In {\em IEEE Symposium on Security \& Privacy (S\&P)}, 2013.

\bibitem{yu2016tracking}
Zhonghao Yu, Sam Macbeth, Konark Modi, and Josep~M. Pujol.
\newblock Tracking the trackers.
\newblock In {\em International Conference on World Wide Web (WWW)}, 2016.

\bibitem{URL_CRITEOAI}
{Web-scale ML : learning is not the (only) point.}
\newblock \url{https://labs.criteo.com/2018/05/ml-model-deployment/}.

\bibitem{URL_GDPR}
{General Data Protection Regulation (GDPR).}
\newblock \url{https://gdpr-info.eu/}.

\bibitem{URL_CCPA}
{California Consumer Privacy Act (CCPA)}.
\newblock \url{https://oag.ca.gov/privacy/ccpa}.

\bibitem{rl}
Richard Sutton and Andrew Barto.
\newblock {\em {Reinforcement Learning: An Introduction}}.
\newblock MIT Press, 1998.

\bibitem{Bashir19adpreferencemanagersndss}
Muhammad~Ahmad Bashir, Umar Farooq, Maryam Shahid, Muhammad~Fareed Zaffar, and
  Christo Wilson.
\newblock Quantity vs. quality: Evaluating user interest profiles using ad
  preference managers.
\newblock In {\em Network and Distributed Systems Security (NDSS) Symposium},
  2019.

\bibitem{urban2019study}
Tobias Urban, Dennis Tatang, Martin Degeling, Thorsten Holz, and Norbert
  Pohlmann.
\newblock {A study on subject data access in online advertising after the
  GDPR}.
\newblock In {\em International Workshop on Data Privacy Management}. 2019.

\bibitem{papadopoulos2017if}
Panagiotis Papadopoulos, Nicolas Kourtellis, Pablo~Rodriguez Rodriguez, and
  Nikolaos Laoutaris.
\newblock If you are not paying for it, you are the product: How much do
  advertisers pay to reach you?
\newblock In {\em Internet Measurement Conference (IMC)}, 2017.

\bibitem{cook20headerbiddingpets}
John Cook, Rishab Nithyanand, and Zubair Shafiq.
\newblock Inferring tracker-advertiser relationships in the online advertising
  ecosystem using header bidding.
\newblock {\em Privacy Enhancing Technologies Symposium (PETS)}, 2020.

\bibitem{URL_ORACLE_WHITEBOOK}
{Oracle Data Cloud Registry 2019 Data Directory}.
\newblock
  \url{https://www.oracle.com/us/solutions/cloud/data-directory-2810741.pdf}.

\bibitem{le2014distributed}
Quoc Le and Tomas Mikolov.
\newblock Distributed representations of sentences and documents.
\newblock In {\em International Conference on Machine Learning (ICML)}, 2014.

\bibitem{patent_ad_targeting}
Taher~H. Haveliwala, Glen~M. Jeh, and Sepandar~D. Kamvar.
\newblock {Targeted advertisements based on user profiles and page profile},
  2012.

\bibitem{patent_ad_targeting_2}
Darrell Anderson, Paul Buchheit, Alexander~Paul Carobus, Yingwei Cui,
  Jeffrey~A. Dean, Georges~R. Harik, Deepak Jindal, and Narayanan Shivakumar.
\newblock {Serving advertisements based on content}, 2006.

\bibitem{openai2017a2c}
{OpenAI Baselines: ACKTR \& A2C.}
\newblock \url{https://openai.com/blog/baselines-acktr-a2c/}.

\bibitem{kim2014convolutional}
Yoon Kim.
\newblock Convolutional neural networks for sentence classification.
\newblock In {\em Conference on Empirical Methods in Natural Language
  Processing ({EMNLP})}, 2014.

\bibitem{hausknecht2015deep}
Matthew Hausknecht and Peter Stone.
\newblock {Deep Recurrent Q-Learning for Partially Observable MDPs}.
\newblock {\em AAAI Conference on Artificial Intelligence (AAAI)}, 2015.

\bibitem{xu2019experience}
Zhiyuan Xu, Jian Tang, Chengxiang Yin, Yanzhi Wang, and Guoliang Xue.
\newblock Experience-driven congestion control: When multi-path tcp meets deep
  reinforcement learning.
\newblock {\em IEEE Journal on Selected Areas in Communications},
  37(6):1325--1336, 2019.

\bibitem{zhou2015c}
Chunting Zhou, Chonglin Sun, Zhiyuan Liu, and Francis Lau.
\newblock A c-lstm neural network for text classification.
\newblock {\em arXiv preprint arXiv:1511.08630}, 2015.

\bibitem{URL_WEBEXTENSION}
{Anatomy of an extension.}
\newblock
  \url{https://developer.mozilla.org/en-US/docs/Mozilla/Add-ons/WebExtensions/Anatomy_of_a_WebExtension}.

\bibitem{pass2006picture}
Greg Pass, Abdur Chowdhury, and Cayley Torgeson.
\newblock A picture of search.
\newblock In {\em Proceedings of the 1st international conference on Scalable
  information systems}, pages 1--es, 2006.

\bibitem{URL_WhoisXMLAPI}
{WhoisXMLAPI for URL categorization.}
\newblock \url{https://whois.whoisxmlapi.com/}.

\bibitem{URL_ALEXASITESBYCATEGORY}
{Alexa - Top Sites by Category: The top 500 sites on the web}.
\newblock \url{https://www.alexa.com/topsites/category}.

\bibitem{gagniuc2017markov}
Paul~A Gagniuc.
\newblock {\em Markov chains: from theory to implementation and
  experimentation}.
\newblock John Wiley \& Sons, 2017.

\bibitem{URL_ADBLOCK}
{EasyList}.
\newblock \url{https://easylist.to/}.

\bibitem{bottou2010large}
L{\'e}on Bottou.
\newblock Large-scale machine learning with stochastic gradient descent.
\newblock In {\em International Conference on Computational Statistics}. 2010.

\bibitem{chanchary2015user}
Farah Chanchary and Sonia Chiasson.
\newblock User perceptions of sharing, advertising, and tracking.
\newblock In {\em Eleventh Symposium On Usable Privacy and Security
  ($\{$SOUPS$\}$ 2015)}, pages 53--67, 2015.

\bibitem{cyphers2019behind}
Bennett Cyphers and Gennie Gebhart.
\newblock {\em {Behind the One-Way Mirror: A Deep Dive Into the Technology of
  Corporate Surveillance}}.
\newblock Electronic Frontier Foundation, 2019.

\bibitem{surveillance2019amnesty}
Amnesty International.
\newblock Surveillance giant: How the business model of google and facebook
  threatens human rights.
\newblock 2019.

\bibitem{amos2020privacy}
Ryan Amos, Gunes Acar, Elena Lucherini, Mihir Kshirsagar, Arvind Narayanan, and
  Jonathan Mayer.
\newblock Privacy policies over time: Curation and analysis of a
  million-document dataset.
\newblock {\em arXiv:2008.09159}, 2020.

\bibitem{vail2008empirical}
Matthew~W Vail, Julia~B Earp, and Annie~I Ant{\'o}n.
\newblock An empirical study of consumer perceptions and comprehension of web
  site privacy policies.
\newblock {\em IEEE Transactions on Engineering Management}, 55(3):442--454,
  2008.

\bibitem{mathur2019dark}
Arunesh Mathur, Gunes Acar, Michael~J Friedman, Elena Lucherini, Jonathan
  Mayer, Marshini Chetty, and Arvind Narayanan.
\newblock Dark patterns at scale: Findings from a crawl of 11k shopping
  websites.
\newblock {\em ACM Conference on Computer-Supported Cooperative Work and Social
  Computing (CSCW)}, 2019.

\bibitem{brunton2013political}
Finn Brunton and Helen Nissenbaum.
\newblock Political and ethical perspectives on data obfuscation.
\newblock {\em Privacy, due process and the computational turn: The philosophy
  of law meets the philosophy of technology}, 2013.

\bibitem{Johnson2019ConsumerPrivacyOpt}
Garrett~A Johnson, Scott~K Shriver, and Shaoyin Du.
\newblock Consumer privacy choice in online advertising: Who opts out and at
  what cost to industry?
\newblock {\em Marketing Science}, 39(1):33--51, 2020.

\bibitem{deepak2019effect}
Deepak Ravichandran and Nitish Korula.
\newblock Effect of disabling third-party cookies on publisher revenue.
\newblock Google, 2019.

\bibitem{URL_PAGEVISIBILITYAPI}
{Mozilla page visibility API}.
\newblock
  \url{https://developer.mozilla.org/en-US/docs/Web/API/Page_Visibility_API}.

\bibitem{URL_CHROMIUMTHROTTLE}
Tab throttling and more performance improvements in {Chrome} {M87}.

\bibitem{laperdrix2021fingerprinting}
Pierre Laperdrix, Oleksii Starov, Quan Chen, Alexandros Kapravelos, and Nick
  Nikiforakis.
\newblock Fingerprinting in style: Detecting browser extensions via injected
  style sheets.
\newblock In {\em 30th $\{$USENIX$\}$ Security Symposium ($\{$USENIX$\}$
  Security 21)}, 2021.

\bibitem{karami2020carnus}
Soroush Karami, Panagiotis Ilia, Konstantinos Solomos, and Jason Polakis.
\newblock Carnus: Exploring the privacy threats of browser extension
  fingerprinting.
\newblock In {\em Proceedings of the Symposium on Network and Distributed
  System Security (NDSS)}, 2020.

\bibitem{starov2017xhound}
Oleksii Starov and Nick Nikiforakis.
\newblock {Xhound: Quantifying the fingerprintability of browser extensions}.
\newblock In {\em 2017 IEEE Symposium on Security and Privacy (SP)}, pages
  941--956. IEEE, 2017.

\bibitem{sjosten2017discovering}
Alexander Sj{\"o}sten, Steven Van~Acker, and Andrei Sabelfeld.
\newblock Discovering browser extensions via web accessible resources.
\newblock In {\em Proceedings of the Seventh ACM on Conference on Data and
  Application Security and Privacy}, pages 329--336, 2017.

\bibitem{URL_ABP}
{Surf The Web With No Annoying Ads}.
\newblock {\url{https://adblockplus.org}}, 2019.

\bibitem{URL_FIREFOX}
{Trackers and scripts Firefox blocks in Enhanced Tracking Protection.}
\newblock \url{https://www.ghostery.com/}.

\bibitem{URL_BRAVE}
{Brave browser.}
\newblock \url{https://brave.com/features/}.

\bibitem{storey2017future}
Grant Storey, Dillon Reisman, Jonathan Mayer, and Arvind Narayanan.
\newblock The future of ad blocking: An analytical framework and new
  techniques.
\newblock {\em arXiv:1705.08568}, 2017.

\bibitem{gugelmann2015automated}
David Gugelmann, Markus Happe, Bernhard Ager, and Vincent Lenders.
\newblock An automated approach for complementing ad blockers’ blacklists.
\newblock {\em Privacy Enhancing Technologies Symposium (PETS)}, 2015.

\bibitem{bau2013promising}
Jason Bau, Jonathan Mayer, Hristo Paskov, and John~C Mitchell.
\newblock A promising direction for web tracking countermeasures.
\newblock {\em Workshop on Web 2.0 Security and Privacy}, 2013.

\bibitem{ikram2017towards}
Muhammad Ikram, Hassan~Jameel Asghar, Mohamed~Ali Kaafar, Anirban Mahanti, and
  Balachandar Krishnamurthy.
\newblock Towards seamless tracking-free web: Improved detection of trackers
  via one-class learning.
\newblock {\em Privacy Enhancing Technologies Symposium (PETS)}, 2017.

\bibitem{shuba2018nomoads}
Anastasia Shuba, Athina Markopoulou, and Zubair Shafiq.
\newblock {NoMoAds: Effective and efficient cross-app mobile ad-blocking}.
\newblock {\em Privacy Enhancing Technologies Symposium (PETS)}, 2018.

\bibitem{wu2016machine}
Qianru Wu, Qixu Liu, Yuqing Zhang, Peng Liu, and Guanxing Wen.
\newblock A machine learning approach for detecting third-party trackers on the
  web.
\newblock In Ioannis Askoxylakis, Sotiris Ioannidis, Sokratis Katsikas, and
  Catherine Meadows, editors, {\em European Symposium on Research in Computer
  Security (ESORICS)}, 2016.

\bibitem{achara2016mytrackingchoices}
Jagdish~Prasad Achara, Javier Parra-Arnau, and Claude Castelluccia.
\newblock Mytrackingchoices: Pacifying the ad-block war by enforcing user
  privacy preferences, 2016.

\bibitem{meng2016trackmeornot}
Wei Meng, Byoungyoung Lee, Xinyu Xing, and Wenke Lee.
\newblock Trackmeornot: Enabling flexible control on web tracking.
\newblock In {\em Proceedings of the 25th International Conference on World
  Wide Web}, WWW '16, page 99–109, Republic and Canton of Geneva, CHE, 2016.
  International World Wide Web Conferences Steering Committee.

\bibitem{mughees2017detecting}
Muhammad~Haris Mughees, Zhiyun Qian, and Zubair Shafiq.
\newblock Detecting anti ad-blockers in the wild.
\newblock {\em Privacy Enhancing Technologies Symposium (PETS)}, 2017.

\bibitem{nithyanand2016adblocking}
Rishab Nithyanand, Sheharbano Khattak, Mobin Javed, Narseo Vallina-Rodriguez,
  Marjan Falahrastegar, Julia~E Powles, Emiliano De~Cristofaro, Hamed Haddadi,
  and Steven~J Murdoch.
\newblock Adblocking and counter blocking: A slice of the arms race.
\newblock In {\em USENIX Workshop on Free and Open Communications on the
  Internet (FOCI)}, 2016.

\bibitem{finnn2015obfuscationbook}
Finn Brunton and Helen Nissenbaum.
\newblock {Obfuscation: A User's Guide for Privacy and Protest}.
\newblock 2015.

\bibitem{xing2013takethis}
Xingyu Xing, Wei Meng, Dan Doozan, Alex~C. Snoeren, Nick Feamster, and Wenke
  Lee.
\newblock Take this personally: Pollution attacks on personalized services.
\newblock In {\em 22nd {USENIX} Security Symposium ({USENIX} Security 13)},
  pages 671--686, Washington, D.C., August 2013. {USENIX} Association.

\bibitem{ming2014pwned}
Wei Meng, Xinyu Xing, Anmol Sheth, Udi Weinsberg, and Wenke Lee.
\newblock Your online interests: Pwned! a pollution attack against targeted
  advertising.
\newblock In {\em Proceedings of the 2014 ACM SIGSAC Conference on Computer and
  Communications Security}, CCS '14, page 129–140, New York, NY, USA, 2014.
  Association for Computing Machinery.

\bibitem{kim2018adbudgetkiller}
I~Luk Kim, Weihang Wang, Yonghwi Kwon, Yunhui Zheng, Yousra Aafer, Weijie Meng,
  and Xiangyu Zhang.
\newblock Adbudgetkiller: Online advertising budget draining attack.
\newblock In {\em Proceedings of the 2018 World Wide Web Conference}, WWW '18,
  page 297–307, Republic and Canton of Geneva, CHE, 2018. International World
  Wide Web Conferences Steering Committee.

\bibitem{chierichetti2012arewebusersmarkovian}
Flavio Chierichetti, Ravi Kumar, Prabhakar Raghavan, and Tamas Sarlos.
\newblock Are web users really markovian?
\newblock In {\em Proceedings of the 21st International Conference on World
  Wide Web}, WWW '12, page 609–618, New York, NY, USA, 2012. Association for
  Computing Machinery.

\end{thebibliography}






%



\end{document}